\documentclass[trsc]{informs3} 

\DoubleSpacedXI

\usepackage{natbib}
 \bibpunct[, ]{(}{)}{,}{a}{}{,}%
 \def\bibfont{\small}%

\usepackage[utf8]{inputenc}
\usepackage[acronym,nomain]{glossaries}
\usepackage{svg}
\usepackage{float}
\usepackage{pifont}
\usepackage{caption}
\usepackage{booktabs}
\usepackage{makecell}
\usepackage{adjustbox}
\usepackage{subcaption}
\usepackage{algorithm}
\usepackage{arydshln}
\usepackage[noend]{algpseudocode}
\usepackage{enumitem}

\usepackage{url}

\usepackage{breakurl}
\usepackage[breaklinks]{hyperref}

\newacronym{diprp}{diPRP}{Dynamic In-store Picker Routing Problem}
\newacronym{mip}{MIP}{Mixed Integer Programming}
\newacronym{mdp}{MDP}{Markov Decision Process}
\newacronym{srp}{SRP}{Shopping Routing Problem}
\newacronym{tsp}{TSP}{Traveling Salesman Problem}

\newacronym{iot}{IoT}{Internet of Things}

\newcommand{\myfigref}[1]{Figure \ref{#1}}
\newcommand{\mytabref}[1]{Table \ref{#1}}
\newcommand{\mysecref}[1]{Section \ref{#1}}
\newcommand{\myalgref}[1]{Algorithm \ref{#1}}

\TheoremsNumberedThrough     

\EquationsNumberedThrough    

\begin{document}




\TITLE{Playing hide and seek: tackling in-store picking operations while improving customer experience}

\ARTICLEAUTHORS{%
\AUTHOR{Fábio Neves-Moreira, Pedro Amorim}
\AFF{\EMAIL{fabio.s.moreira@inesctec.pt},  \EMAIL{pamorim@fe.up.pt} \\ 
INESC TEC, Faculty of Engineering, University of Porto, Porto 4200–465, Portugal
}
} 






\ABSTRACT{%

	The evolution of the retail business presents new challenges and raises pivotal questions on how to reinvent stores and supply chains to meet the growing demand of the online channel. One of the recent measures adopted by omnichannel retailers is to address the growth of online sales using in-store picking, which allows serving online orders using existing assets. However, it comes with the downside of harming the offline customer experience. To achieve picking policies adapted to the dynamic customer flows of a retail store, we formalize a new problem called \gls{diprp}.
	In this relevant problem -- \gls{diprp} -- a picker tries to pick online orders while minimizing customer encounters. We model the problem as a \gls{mdp} and solve it using a hybrid solution approach comprising mathematical programming and reinforcement learning components. Computational experiments on synthetic instances suggest that the algorithm converges to efficient policies. Furthermore, we apply our approach in the context of a large European retailer to assess the results of the proposed policies regarding the number of orders picked and customers encountered.
	Our work suggests that retailers should be able to scale the in-store picking of online orders without jeopardizing the experience of offline customers. The policies learned using the proposed solution approach reduced the number of customer encounters by more than $50\%$ when compared to policies solely focused on picking orders. Thus, to pursue omnichannel strategies that adequately trade-off operational efficiency and customer experience, retailers cannot rely on actual simplistic picking strategies, such as choosing the shortest possible route.
	
}




\maketitle

%



\glsresetall
\section{Introduction}
\label{sec:introduction}

The future of retail is very uncertain. This uncertainty is strongly connected to the pace at which customer consumption patterns, channel shifts, and expectations around speed and convenience are changing. The Covid-19 pandemic fuelled this unpredictable behavior, but these changes are here to stay. At the moment, several pivotal questions posed by investors, entrepreneurs, business professionals, and academics connected to retail remain unanswered \citep{caro2020}.

One of the most critical challenges faced by large retailers is related to the future of physical retail stores \citep{mckinsey2022}. While some believe that physical retail may be ``dead'', others have many reasons to defend their existence as a way to differentiate from competition, offering remarkable experiential shopping based on good customer service, exuberant stores, and online-to-physical channels harmonization \citep{Forbes2018}. Nowadays, many retailers recognize that focusing on omnichannel retailing is the way to go as a means for enriching customer value proposition and improving operational efficiency \citep{gao2017}. In particular, the physical channel still remains important for a substantial part of grocery retail customers, who find it convenient to place an order online and have their groceries picked up at a nearby retail location \citep{chloe2019, vyt2022}. However, many challenges emerge from this offline-online dichotomy. For example, it is not clear how to provide the online level of information to offline customers \citep{fei2017}, to which channel should the most stock be committed \citep{jia2021}, and which assortment decisions should be made in the offline and online channels \citep{chen2021}. 

What is clear is that physical retail stores offer tremendous value in terms of convenience and speed, particularly in food retail \citep{jayakumar2021}. For example, these assets support the recent trend in omnichannel retailing to shift from e-commerce to quick commerce (Q-Commerce), where the focus is on providing small quantities of goods ranging from groceries, stationeries to over-the-counter medicines in short delivery times of $10$ to $30$ minutes. This type of service, sometimes called on-demand delivery, requires faster fulfillment strategies that require a shift from traditional warehouses located on the outskirts to micro-warehouses or physical retail stores located near the delivery locations. This means that the number of in-house and third-party pickers inside retail stores should increase considerably in the upcoming times, so that retailers can face the growth in the number of online and pick-in-store orders that need to be delivered in short delivery times. These omnichannel fulfillment strategies should be further propelled when fully digitized stores are deployed, allowing retailers to profit from recent technologies, such as \gls{iot}, advanced robotics, and digital twins \citep{olsen2020}.




Motivated by the real case of a European retailer, this paper will focus on the satisfaction of online orders that have to be picked by in-store pickers. These orders are posted on the website of the retailer and can be delivered at home or picked in-store by the customer. This context is a very challenging one, and aiding pickers during their picking routes is becoming more and more important due to four main reasons. First, the online business is drastically increasing in recent times, as customers have been almost forced to experiment online shopping \citep{covidshift}. Second, most physical stores do not have sufficient backroom space to settle a warehouse optimized for picking all products that can appear in online orders \citep{pires2021}. Third, since more and more third-party pickers (i.e., Uber, Glovo, and Delivery Hero) are now shopping in the stores due to the recent interest in Q-Commerce \citep{qcommerce}, it is important to provide a service that can aid them to navigate through retail stores when they are not properly aware of the locations of every product. Fourth, and more importantly, with the referred influx of internal and third-party pickers, the in-store customer experience is at risk, conflicting with online-offline strategies followed by grocery retailers.

In this paper, to achieve picking policies that can be adapted to the dynamic customer flows of a retail store and help to mitigate the aforementioned challenges, we introduce a new problem called \gls{diprp}. This dynamic problem considers a dynamic retail store environment and consists in making decisions so that a picker maximizes the number of orders picked while minimizing the number of customer encounters. Note that modeling uncertain customer flows inside a store is a core element of this challenge. Traditional methods for aiding pickers based on stochastic mathematical programming are likely to result in intractable \gls{mip} models and are not suited to capture the dynamics of a sequential decision problem. Therefore, the \gls{diprp} is modeled as a \gls{mdp} and solved through a well-known reinforcement learning technique. The picking policies are obtained employing a hybrid Q-learning algorithm that maximizes immediate and future rewards (i.e., a positive reward is received when a product is picked and a negative reward one is received when a customer is encountered) by providing the best actions (i.e., picker movements) to be performed by the picker. The objective is to pick the largest number of orders (operational efficiency) while avoiding customer encounters (shopping experience).

We provide evidence on how one can profit from real-world data and orchestrate mathematical programming, simulation, and machine learning techniques to solve a relevant practical problem in omnichannel retail - the \gls{diprp}. The scientific contributions of this work are threefold:
\begin{enumerate}
    \item We formalize a new and relevant problem in retail where a picker picks online and pick-in-store orders inside a retail store, while avoiding physical customer encounters - the \gls{diprp}.
    \item We solve the dynamic problem using a hybrid Q-learning algorithm, which includes \gls{mip} and \gls{mdp} components to determine a picking policy that is adapted to the dynamic environment of an in-store picking operation.
    \item We provide computational experiments on the algorithm convergence using synthetic instances and derive managerial insights on four different picking policies that were validated on a real-world application partnered with a large European retailer. These insights aim at helping retailers to make better decisions related to their omnichannel strategies. Namely, proving a solid ground to scale new fulfillment options.
\end{enumerate}
The remainder of this paper is organized as follows. Section \ref{sec:literature} reviews the relevant literature related to in-store picker routing problems.
In Section \ref{sec:problemdescription} we describe the problem and present the inherent retail store simulation environment to be used in our study. 
Section \ref{sec:solution_approach} details the proposed solution approach based on a Q-learning algorithm. Section \ref{sec:computational_experiments} presents the computational experiments performed on synthetic instances, which showcases the algorithm convergence. 
Section \ref{sec:real_world_application} exposes a real-world application in a European retail store.
Finally, in Section \ref{sec:conclusions}, we present the main conclusions of this work and suggest future research directions.

\section{Literature Review}
\label{sec:literature}
In the following subsections, we review recent approaches dealing with picker routing problems (\mysecref{subsec:picker_routing_problems}), dynamic routing problems \mysecref{subsec:dynamic_routing_problem}, and dynamic picker routing problems \mysecref{subsec:dynamic_picker_routing_problem}. We consider that these topics cover the modeling and solution approach techniques needed to tackle our problem - \gls{diprp} -, which consists of picking products in a dynamic store environment, a new type of picking operation.

\subsection{Picker routing problems}
\label{subsec:picker_routing_problems}

The picker or order routing problem was introduced by \citet{ratliff1983} and is typically modeled as a \gls{tsp} on a Steiner graph \citep{letchford2013, cambazard2018, rodriguezpereira2019}. Several problem extensions have been proposed in the literature to adapt the standard model to real-world cases with specific business constraints \citep{chabot2017, sadia2018, ardjmand2018} and to combine subsequent problems, such as storage and batching \citep{vangils2018}. In terms of solution methods, this problem is usually tackled with heuristics \citep{theys2010}, but important problem properties were discovered to increase the efficiency of exact methods based on mathematical formulations \citep{cornujols1985, pansart2018}. Recently, new formulations have been proposed to increase the size of the instances solved exactly \citep{scholz2016} and to integrate the order picking problem with other activities in the supply chain \citep{roodbergen2015}, as this remains to be a relevant problem in warehouse operations management. 
Picker routing problems are typically solved within narrow aisle warehouses \citep{roodbergen2001a, roodbergen2001b, dekoster2007, chabot2018, masae2020} and not inside stores. Considering layouts that are not optimized for picking operations and can be crowded with customers lead to a new context that has never been studied in the order picking literature. In a store, product layouts may be very irregular and most graph properties assumed in order picking literature are not applicable. Furthermore, customer shopping paths can be very chaotic and are not known \textit{apriori}, meaning that picking problems are \textit{more} stochastic and dynamic.

\subsection{Dynamic routing problems}
\label{subsec:dynamic_routing_problem}

Although stochastic dynamic routing has a 30-year history, the majority of the papers published in this field are recent. More than half of the papers in the literature review of \citet{ulmer2020a} have been published after 2010. 
Other literature reviews on this topic have been published by \cite{berbeglia2010}, \cite{pillac2013}, \cite{ritzinger2016}, \cite{psaraftis2016}, and  \cite{ojedarios2021}. It is worth mentioning that multiple trends (e.g., e-commerce growth, sharing economies, and sustainability) and technological advances (e.g., digital connectivity, big data, and automation) are fostering the development of new approaches to tackle dynamic routing problems \citep{savelsbergh2016}. To solve these problems, the unified framework for stochastic programming proposed by \citet{powell2019} has been determinant, as researchers have been adapting it to tackle dynamic problems modeled as \glspl{mdp}. To turn dynamic programming applicable to large real-world problems, the approximate dynamic programming approaches presented in \citet{powell2007} have also been a preponderant reference in this field.

More recently, several attempts have also been tried to solve routing problems modeled as \glspl{mdp} through Reinforcement Learning techniques (\cite{nazari2018}, \cite{jingwen2021}, \cite{kullman2022}). However, despite their potential to be used in new contexts, these methods are still seen as unexplored by the routing community, as suggested by \cite{hildebrandt2021}.


\subsection{Dynamic picker routing problem}
\label{subsec:dynamic_picker_routing_problem}

In some business contexts, it is interesting to consider a dynamic variant of the picker routing problem. These problems arise when new information allows for better planning or execution of the picking operations (i.e. a new order arrived). With respect to dynamic picker routing problems, the literature is very scarce. 
\cite{yeming2008} consider picking operations where picking information can dynamically change in a picking cycle. Based on a stochastic polling theory approach, the authors suggest that shorter order throughput times and higher on-time service completion ratios can be achieved. \cite{lu2016} consider dynamic order-picking strategies that allow for changes in the picking lists during a picking route. The authors allow for the re-optimization of the optimal route during the picking operation and achieve better solution quality when compared to the static version of the problem. 

Note that these authors consider that dynamic information arriving is only related to customer arrivals. However, in a store, there can be two types of entities performing picking, the offline customers and the online order pickers. Pickers can gather information on the location of the customers as they perform their picking paths. This maps into a new dynamic picking routing problem. However, the interactions between in-store pickers and in-store customers have not been considered yet, that is, no approach has been proposed for solving the dynamic in-store picker routing problem.

Despite the considerable number of articles dealing with picker routing problems, no publication has considered the new environment that is faced by in-store picking teams. Indeed, the pandemic context of 2020 brought new operations to be managed and new challenges to be tackled while pursuing different objectives. For instance, constraints imposed on the number of customers inside of a store may turn retailers interested in guiding customers through their shopping paths to accelerate customers' shopping process, as opposed to making them stay for longer inside the store \citep{hui2013, boros2015, hipara2021}. Each customer in-store is also solving its picker routing problem to define its shopping path. 
The fact that customers do not know the exact position of the products and that some customers may have picking sequence preferences \citep{larson2005} can induce chaotic flows of customers through the aisles of a store \citep{chen2015, li2017}. This means that the state of the store is not known \textit{apriori} when the picking route of each picker is being defined, it is stochastic. Therefore, in case we are interested in considering picking paths that do not disturb customers’ shopping experience, the picker routing problem proposed approaches fall short on tackling this issue. This is a gap that we address in this work.

\section{Problem description}
\label{sec:problemdescription}
\def \myPickingPositionIndex         { i }
\def \myPickingPositionIndexb        { j }
\def \myProductIndex                 { p }

\def \myGraph 							{\mathcal G} 
\def \myNodes 							{\mathcal V} 
\def \myEdges 							{\mathcal E} 
\def \myPickingPositions 				{\mathcal S} 

\def \myEdgeWeight      { w_{\myPickingPositionIndex\myPickingPositionIndexb} }
\def \myProductWeight   { w_{\myProductIndex} }
\def \myProductDeadline { r_{\myProductIndex} }

\def \myDVX { x_{ \myPickingPositionIndex \myPickingPositionIndexb } }

\def \myDVL { l_{ \myPickingPositionIndex \myPickingPositionIndexb } } 
\def \myDVY { y_{ \myPickingPositionIndex } } \

In this section, we formalize the \gls{diprp} by describing all the relevant elements included in the considered retail store environment, the picker agents that will make decisions inside the store, and the \gls{mdp} that is proposed to model the associated dynamic problem. 
Hence, \mysecref{subsec:environment} details the retail store environment, including the store graph, the models for simulating the shopping paths of physical customers, and the online order arrivals composed of a set of picking positions. \mysecref{subsec:picker_agents} clarifies the behavior of a picker considering the information that is available in each step (e.g., next product to be picked and physical customers nearby). \mysecref{subsec:mdp} describes the \gls{mdp} that models the dynamic problem we aim to solve, including decision epochs, states, actions, transition function, rewards, and objective function.

\subsection{Retail Store Environment}
\label{subsec:environment}

\def \myPhysicalCustomersIndex  { c }
\def \myOnlineOrderIndex        { o }
\def \myEpoch                   { k }

\def \myPhysicalCustomers       { \mathcal C }
\def \myOnlineOrders            { \mathcal O }

\def \myPhysicalCustomersPath           { \mathcal \theta_{\myPhysicalCustomersIndex} }
\def \myOnlineOrderShoppingList         { \mathcal L_{\myOnlineOrderIndex} }
\def \myOnlineOrderPickingSequence      { \mathcal \theta_{\myOnlineOrderIndex} }

\def \myOnlineOrderRate     { \lambda^{Online} }
\def \myStoreCustomerRate   { \lambda^{Store}  }

\def \myStoreCustomerSpeed  {speed^{Customer} }
\def \myStoreCustomerPickingTime    {service\_time^{Customer} }

\def \myPickerSpeed         {speed^{Picker} }
\def \myPickerPickingTime   {service\_time^{Picker} }

\def \myStoreMaxPhysicalCustomers       { |\mathcal C| } 
\def \myNodeMaxPhysicalCustomers        { m } 

To model a realistic retail store environment, we resort to a simulation environment considering physical customer arrivals and online order arrivals. The store is modeled as a sparse graph $\myGraph = (\myNodes, \myEdges)$ where the set of vertices (locations in-store) is partitioned into vertex $0$, which is the entrance of the store where the shopping paths start, vertices $\{1, ..., v\}$, corresponding to $v$ positions inside the store, and vertex $v+1$, which is the position where the shopping paths end (i.e., a cashier or an order preparation zone).
Physical customers arrive at the store following a Poisson distribution with a mean value of $\myStoreCustomerRate$ customers per time period. Each physical customer $\myPhysicalCustomersIndex \in \myPhysicalCustomers$ moves through the aisles of the store navigating through the graph $\myGraph$ at a speed $\myStoreCustomerSpeed$ and according to a given shopping path $\myPhysicalCustomersPath$ containing a sequence of locations in-store. A physical customer takes $\myStoreCustomerPickingTime$ time periods to pick a product located at a certain node. The maximum number of customers inside the store can never surpass the capacity of the store $\myStoreMaxPhysicalCustomers$ and the maximum number of customers in the same store node is $\myNodeMaxPhysicalCustomers$.
The arrival rate of online orders is also a random variable, which follows a Poisson distribution with a mean value of $\myOnlineOrderRate$ orders per time period. Each online order $\myOnlineOrderIndex \in \myOnlineOrders$ is composed of a set of picking positions $\myOnlineOrderShoppingList \subset \myNodes$ that need to be visited to pick the ordered products.

\subsection{Picker agents}
\label{subsec:picker_agents}

Whenever an online order $\myOnlineOrderIndex$ arrives it needs to be allocated to an idle picker. At that moment, a picking sequence $\myOnlineOrderPickingSequence$ is defined according to a business decision rule (e.g., shortest route, heavy items first, and frozen products at the end). In this work, we start by assuming that these picking sequences are minimizing the traveled distance, i.e., shortest route, - maximizing operational efficiency, which is the common case in practice. The picker follows a given picking sequence, yet he is interested in avoiding customer encounters not to disturb their shopping experience. Therefore, between every pair of products to be picked, the picker solves a shortest path problem, while avoiding customers currently in the store. The dynamic component of this problem corresponds to solving a series of dynamic shortest path problems. The fundamental trade-off that is inherent to this problem comes from the fact that the picker follows a predefined sequence of picking positions that is provided by a decision rule adopted by the retail store (e.g., shortest routes), but he/she needs to avoid customer encounters in the path between two picking positions. The position of the customers in-store is not known in advance and it is only revealed when the picker looks through the aisle. Throughout the day, each picker receives several online orders, and thus, several picking sequences to execute. 


\subsection{Dynamic picker routing problem as a Markov decision process}
\label{subsec:mdp}

\def \myTimeHorizon         { T }
\def \myInfo                { \Omega }

\def \myStates              { \mathcal S }
\def \myActions             { \mathcal A }
\def \myObservations        { O }
\def \myRewards             { R }
\def \myTransitions         { P }

\def \myState               { s_{\myEpoch} }
\def \myInitialState               { s_{0} }
\def \myAction              { a_{\myEpoch} }
\def \myCurrentLocation     { n_{\myEpoch} }
\def \myArrivalTime         { t_{\myEpoch} }
\def \myTargetLocation      { z_{\myEpoch} }
\def \myReward              { R_{\myEpoch} }
\def \myExpectedReward      { \widehat{R}_{\myEpoch} }

\def \myRealizationInfo     { \omega }
\def \myNumberOfCustomers   { \phi}

\def \myPolicy              { \pi(s) }
\def \myValueFuction        { V_{\myPolicy}(s) }
\def \myDiscountRate        { \gamma }

To formulate the \gls{diprp} as finite horizon \gls{mdp} with $\myTimeHorizon$ time periods, consider a decision epoch $\myEpoch \in {0,1,.., \myTimeHorizon}$ at which a decision for the next position node to visit inside the store should be made. A new decision epoch $\myEpoch$ is triggered whenever the picker arrives at a new position. Each decision epoch $\myEpoch$ is associated with a decision state $\myState$. At the initial state $s_{0}$ and initial decision epoch $\myEpoch = 0$, the picker is located at the starting depot node $0$. At the terminal decision epoch $\myEpoch = \myTimeHorizon$, the picker returns to the finishing depot node $v+1$ after picking the order that is currently being picked. At a state $\myState$, the system receives updated information on the number of customers at the same position of the picker and adjacent positions (connected by an arc). Based on the system information about current online orders, an action $\myAction$ (i.e., the next position to visit in targeting a product or the depot) needs to be decided. Each position that is visited increases the total travel time and, potentially, the number of customer encounters. The goal of the picker is to find optimal paths, short and quick to traverse while minimizing the number of customer encounters. The picker is positively rewarded for each product picked. Overall, the problem faced by the in-store picker can be modeled as a \gls{mdp} composed of four components, namely, 
(1) a set of states $\myStates$; 
(2) a set of actions $\myActions$; 
(3) a reward function $\myRewards$; 
and the transition probabilities $\myTransitions$.

\subsubsection{Decision epochs}
\label{subsubsec:epochs}
A decision epoch $\myEpoch$ begins when a picker arrives at a new location in-store.

\subsubsection{States}
\label{subsubsec:states}
The state of the system at decision epoch $\myEpoch$ is defined by the tuple $\myState = (\myCurrentLocation, \myTargetLocation, \myArrivalTime)$, where $\myCurrentLocation \in \myNodes$ is the picker's current location, $\myTargetLocation = (\myTargetLocation(1), \myTargetLocation(2),..., \myTargetLocation(|\myOnlineOrderShoppingList|))$ is a vector containing the status of each picking location of an online order $\myOnlineOrderIndex$ at decision epoch $\myEpoch$, and $\myArrivalTime \in [0, \myTimeHorizon]$ is the arrival time at location $\myCurrentLocation$. For each picking location of an online order, $\myTargetLocation(.)$ takes on a value in the set $\{0, 1\}$:
\begin{equation}
\myTargetLocation(.) =
  \begin{cases}
    0, & \text{if picking location has not been visited by time $\myArrivalTime$} \\
    1, & \text{if picking location has been visited by time $\myArrivalTime$}
  \end{cases}
\end{equation}

In the initial state $s_{0} = (0, z_{0}, 0)$, the picker is located at the order preparation zone and $z_{0}(n)$ is $0$ or $1$ for each picking location $n \in \myNodes \setminus \{0\}$. The final decision epoch is attained when the picker reaches the terminal state $s_{K}$ in the set $\{(0, \myTargetLocation, \myArrivalTime): \myArrivalTime \in [0, \myTimeHorizon], \myTargetLocation \in \{0, 1\}^{|\myOnlineOrderShoppingList|} \}$, where the picker has returned to the preparation zone by time $\myTimeHorizon$ and the status of the picking locations of the online order being served is 0 or 1. The state space is the set $\myStates =  \myNodes \times [0, \myTimeHorizon] \times \{0,1\}^{|\myOnlineOrderShoppingList|}$.

\subsubsection{Actions}
An action at decision epoch $\myEpoch$ is an assignment of the picker to an in-store location belonging to $\myNodes$. When the picker is at state $\myState$, the set of feasible actions, corresponding to the possible arcs leaving $\myCurrentLocation$ is given by 
\begin{equation}
    \myActions(\myState) = \{\myAction \in \{n \in \myNodes: (\myState, n) \in \myEdges\} \}:
\end{equation}

\subsubsection{Transition}
\label{subsubsec:transition}
Following the selection of action $\myAction$ from state $\myState$, the process transitions to a new state $s_{\myEpoch+1}$ in decision epoch $\myEpoch+1$. There is no uncertainty regarding the position in which the picker will be in the next decision epoch. However, the unknown information regarding the number of customers encounter is revealed. This information is a realization of $\myInfo_{\myEpoch+1}$  and is given by $\myRealizationInfo_{\myEpoch+1}$. Let $\widehat{R}_{\myEpoch+1}(\myState, \myAction, \myRealizationInfo_{\myEpoch+1})$ be the random negative reward accrued at decision epoch $\myEpoch$ when selecting action $\myAction$ from state $\myState$  and observing random information $\myRealizationInfo_{\myEpoch+1}$.

\subsubsection{Rewards}
\label{subsubsec:rewards}
When the picker occupies state $\myState$ and performs an action $\myAction \in \myActions$, a reward is accrued depending on the conditions of the in-store location to which the picker moves. The reward is composed of four components, (1) fixed negative reward for each picker step, (2) number of in-store customers in the same location of the picker, (3) number of in-store customers in the locations reachable from the picker location, and (4) a positive reward for a picked product. The number of steps given from the last decision epoch is given by $\phi_{1}$, the number of customers in the same location is given by $\phi_{2}$, the number of customers nearby is given by $\phi_{3}$, and the number products picked is given by $\phi_{4}$. These four components can be adjusted using weights $w_1$ to $w_4$, respectively. The reward $R_{\myEpoch}(\myState, \myAction, \myRealizationInfo_{\myEpoch+1})$, which is a stochastic variable depending on the store environment, is defined as
\begin{equation}
\myReward(\myState, \myAction, \myRealizationInfo_{\myEpoch+1}) 
= 
- w_{1} \thinspace \phi_{1}(\myRealizationInfo_{\myEpoch+1})
- w_{2} \thinspace \phi_{2}(\myRealizationInfo_{\myEpoch+1})
- w_{3} \thinspace \phi_{3}(\myRealizationInfo_{\myEpoch+1})
+ w_{4} \thinspace \phi_{4}(\myRealizationInfo_{\myEpoch+1})
\end{equation}

\subsubsection{Objective}
The objective is to maximize the expected sum of rewards across decision epochs, that is, if weights are properly set, trying to pick as much orders as possible while minimizing in-store customer encounters. Let $\pi$ be a sequence of actions ${a_{0}, a_{1}, ..., a_{\myTimeHorizon}}$ for every decision epoch $\myEpoch = 0, ..., \myTimeHorizon$. The aim is to find an optimal policy $\pi\*$ to maximize the total expected reward. The objective function, conditional on initial state $\myInitialState$, is the following
\begin{equation}
\min_{\pi} \mathbb{E}\big[ \sum_{\myEpoch = 0}^{\myTimeHorizon} \myReward | \myInitialState \big]
\end{equation}

\section{Solution approach}
\label{sec:solution_approach}

\begin{figure}[hbt]
	\centering															
	\includegraphics[width=0.55\textwidth,trim = 0mm 35mm 130mm 0mm, clip]{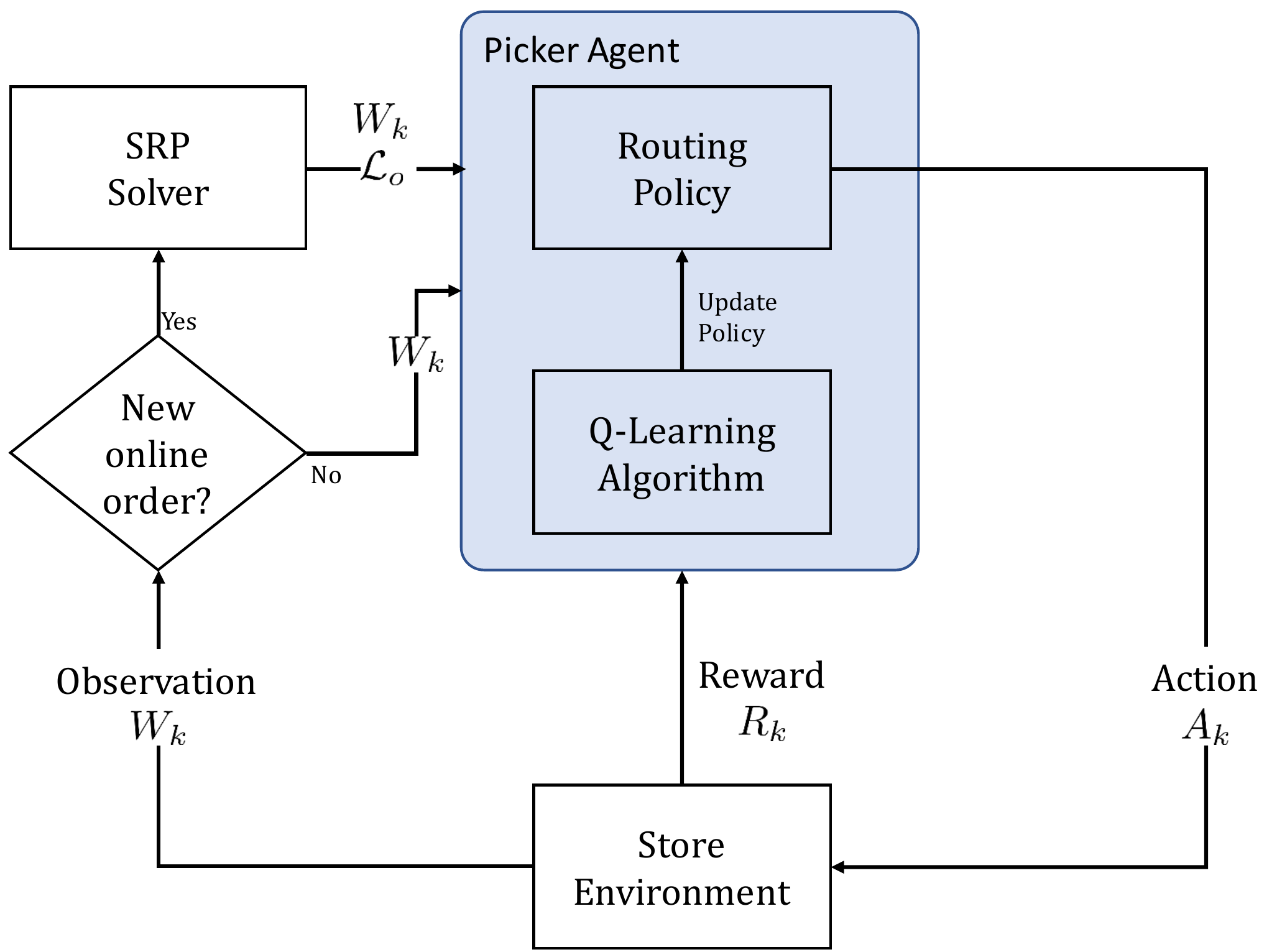}
	\caption{
	Approach overview including a detailed breakdown of the picker agent. 
	}
	\label{fig:approach_overview}
\end{figure}

The \gls{diprp} defined in \mysecref{sec:problemdescription} demands solution techniques with different scopes and purposes: 
(1) whenever an online order arrives, a \gls{srp} model, an \gls{mip}, needs to be solved to obtain a picking sequence according to predefined decision rules;
and (2) for each picking sequence and for every pair of products on that sequence, a sequential decision problem is solved and recursively updates the policy via a Q-learning algorithm. \myfigref{fig:approach_overview} presents an overview of the devised approach, showing where each technique is needed and how the picker agent interacts with the store environment.

In \mysecref{subsubsec:srp_problem_definition}, we describe the problem, the mathematical formulation and the solution approach that are used to obtain the picking sequences to feed the picker agent related procedures.
In \mysecref{subsec:q_learning}, we describe the proposed Q-learning approach to learn an optimal policy for efficiently picking orders while avoiding in-store customer encounters. Section \ref{subsection:illustrative_example} provides an example of picking paths obtained with different policies and an example of what a policy obtained with Q-Learning looks like.

\subsection{\gls{srp} definition}
\label{subsubsec:srp_problem_definition}
To introduce the standard \gls{srp}, consider the retail store environment described in \mysecref{subsec:environment}. Each edge $(\myPickingPositionIndex, \myPickingPositionIndexb) \in \myEdges$ is a pair of positions associated with a sequence of grid steps corresponding to the shortest path between picking position $\myPickingPositionIndex$ and $\myPickingPositionIndexb$. Edges are weighted by a factor $\myEdgeWeight$, typically representing the distance or the duration of the arc. Again, let $\myOnlineOrderShoppingList \subset \myNodes$ be a subset of vertices corresponding to the picking positions that have to be visited to pick a given shopping list $\myOnlineOrderIndex$. The main objective of the picker is to pick all the products in the shopping list while minimizing the distance (or duration) of its shopping path. 

\subsubsection{\gls{srp} Mathematical formulation}
\label{subsubsec:srp_mathematical_formulation}
To model the \gls{srp} we use binary decision variables $\myDVX$ for defining routing decisions. Let $\myDVX$ be the binary variables indicating whether an edge $(\myPickingPositionIndex, \myPickingPositionIndexb)$ is traversed. The proposed formulation reads as follows:

\noindent \gls{srp}:  
\begin{equation}\label{obj}
 \text{minimize} \sum_{(\myPickingPositionIndex, \myPickingPositionIndexb) \in \myEdges}
 \thinspace \myEdgeWeight \thinspace \myDVX
\end{equation}
s.t.
\begin{equation}\label{eqa}
	\sum_{\myPickingPositionIndex \in \myNodes} x_{\myPickingPositionIndexb\myPickingPositionIndex} = 1 
	\quad \forall \thinspace 
	\myPickingPositionIndexb \in \myOnlineOrderShoppingList \cup \{0\}
\end{equation}
\begin{equation}\label{eqb}
	\sum_{\myPickingPositionIndex \in \myNodes} \myDVX = 1 
	\quad \forall \thinspace 
	\myPickingPositionIndexb \in \myOnlineOrderShoppingList \cup \{v+1\}
\end{equation}
\begin{equation}\label{eqc}
	\sum_{\myPickingPositionIndex \in {\mathcal O}}
	\sum_{\myPickingPositionIndexb \in {\mathcal O}}
	\myDVX \leq |\mathcal O|-1  
	\quad \forall \thinspace 
	{\mathcal O} \subset \myNodes, i \neq j
\end{equation}
\begin{equation}\label{eqd}
	x \in \{0, 1\}.
\end{equation}
Objective function (\ref{obj}) minimizes the total distance (or duration) of the shopping path. The flow conservation at the starting position, product picking positions, and finishing position are ensured by constraints (\ref{eqa}) and (\ref{eqb}). Constraints (\ref{eqc}) are added to eliminate sub-tours. Finally, constraints (\ref{eqd}) define the type and bounds of each variable.

\subsubsection{\gls{srp} Solution approach}
\label{subsubsec:srp_solution_appraoch}

In the context of \glspl{diprp}, each \gls{srp} needs to be solved in fractions of a second. Furthermore, in a simulation procedure (later used for optimizing the picking policies), we need to solve hundreds if not thousands of \gls{srp} instances. Relying solely on a mathematical solver is not sufficient when the picking lists comprise a considerably large number of products. For that reason, we tackle this problem with an efficient exact solution approach that can quickly solve the mathematical formulation presented in the previous section. We employ a cutting planes algorithm where cuts are added to remove fractional solutions from the admissible region of the linear relaxation (\myalgref{algo:CuttingPlanes}). 

\def \CutsAdded { CutsAdded }
\begin{algorithm}[hbt]
\caption{Cutting Planes}\label{algo:CuttingPlanes}
\begin{algorithmic}[1]
\footnotesize{
\Procedure{CuttingPlanes($\myNodes$, $\myOnlineOrderShoppingList$)}{}
\State $\CutsAdded \gets True; X \gets \emptyset; Y \gets \emptyset; Z \gets 0;$
    \While{$\CutsAdded$}
        \State $X \gets \text{Solve assignment model (\ref{obj})-(\ref{eqb}) considering sets } \myNodes \text{ and } \myOnlineOrderShoppingList;$ \label{algoline:solve_relaxed_model}
        \State $Y \gets \text{Separate arcs in sets of connected components;}$ \label{algoline:separate_connected_components}
        \State $Z \gets \text{Count the number of connected component sets in Y;}$
        \label{algoline:count_connected_components}
    	\If {$Z = 1$}
        	\State $\CutsAdded \gets False;$ 
    	\Else
    	    \State $\text{Add cuts (\ref{eqc}) to eliminate sub-tours;}$ \label{algoline:has_subtours}
    	    \State $\CutsAdded \gets True;$ \label{algoline:add_cuts}
    	\EndIf
        \If {$\CutsAdded$}
            \State $X \gets \text{Convert positive variables to integer;}$ \label{algoline:convert_to_integer}
    	\Else
        	\If {$\text{All edges in the solution are integer}$}
        	    \State $\myOnlineOrderPickingSequence \gets \text{The picking sequence is obtained from the model solution X;}$
        	    \State \Return $\myOnlineOrderPickingSequence$. \label{algoline:return_optimal_solution}
        	\EndIf   
    	\EndIf
    \EndWhile
\EndProcedure
}
\end{algorithmic}
\end{algorithm}

The algorithm starts by solving a relaxed assignment model (\myalgref{algo:CuttingPlanes}, line \ref{algoline:solve_relaxed_model}) and separating connected component sets (\myalgref{algo:CuttingPlanes}, line \ref{algoline:separate_connected_components}). If there are more than one connected component set, the solution has sub-tours (\myalgref{algo:CuttingPlanes}, line \ref{algoline:has_subtours}) and cuts are added to eliminate these sub-tours (\myalgref{algo:CuttingPlanes}, line \ref{algoline:add_cuts}).
When cuts are added, the edge variables with a positive value are converted to an integer. This process is repeated until no cuts are added after solving the relaxed assignment model and all the edge variables are integer, meaning that the optimal solution is found (\myalgref{algo:CuttingPlanes}, line \ref{algoline:return_optimal_solution}).

\subsection{Q-Learning}
\label{subsec:q_learning} 
The cutting plane approach presented in \mysecref{subsubsec:srp_solution_appraoch} provides a sequence $\myOnlineOrderPickingSequence$ for picking the list of products of online order $\myOnlineOrderIndex$. However, an in-store picker has multiple options to go from one picking position to another. In the \gls{srp} formulation we assume shortest paths, that is, the steps to perform when traversing an arc $(\myPickingPositionIndex, \myPickingPositionIndexb)$ are the ones leading to the shortest distance between $\myPickingPositionIndex$ and $\myPickingPositionIndexb$ (again, note that other decision rules can be used). Therefore, in order not to disturb in-store customers, it may be beneficial to perform different steps or detours while traveling from $\myPickingPositionIndex$ to $\myPickingPositionIndexb$, depending on the unknown patterns related to the movement of the customers in-store.

To obtain a routing policy that follows a given picking sequence but tries to avoid customers, we devise a Q-Learning \citep{watkins1989} approach to solve the \gls{mdp} previously presented in \mysecref{subsec:mdp}. This technique belongs to reinforcement learning, which is a branch of approximate dynamic programming. In simple terms, it is a value-based method which searches for action-value functions $Q(s_k, a)$ representing the future expected reward of taking action $a$ from state $s_k$.

Let us denote $V_\pi(\myState)$ as the value function that gives the total expected reward if we start from a state $\myState$ and follow policy $\pi$. This value includes the immediate reward received at decision epoch $\myEpoch+1$ and the expected rewards to obtain in the future until the final decision epoch $\myTimeHorizon$. Our objective is to find the action array that maximizes the total reward obtained through the entire process. Considering the Bellman Equation \citep{powell2007},  $V^{*}(\myState) = \max\limits_{\myAction \in \myActions(\myAction)} \{R_{\myEpoch+1} + {\mathbb{E}} [ V(s_{\myEpoch+1}) | \myState, a] \}$, an optimal action $\myAction^{\*}$ is implemented by solving the following expression, 
\begin{equation}
    \myAction^{\*} = \argmax\limits_{\myAction \in \myActions(\myState)} \{R_{\myEpoch+1} + {\mathbb{E}} [ V(s_{\myEpoch+1}) | \myState, a] \}
\end{equation}
In the problem we are considering, agents do not control the characteristics of the state they will end up in due to the uncertain customer shopping paths. Therefore, they only have limited influence in the next state that they will visit, depending on their current state $\myState$ and a performed action $\myAction$. Let $Q_{\pi}(\myState, \myAction)$ be a function (Q-function) representing the quality of action $\myAction$ when the agent is at a given state $\myState$ and follows policy $\pi$. We can define the optimal Q-function as  $Q^{*}(\myState, \myAction)$. The relation between $V^{*}(\myState)$ and $Q^{*}(\myState, \myAction)$ is given by the following expression,
\begin{equation}
    V^{*}(\myState) = \max\limits_{a \in \myActions(\myState)} Q^{*}(\myState, a)
\end{equation}
Therefore, an optimal policy $\pi^{*}$ for the problem we aim to solve can be derived through the following expression,
\begin{equation}
    \pi^{*}(\myState) = \argmax\limits_{a \in \myActions(\myState)} Q^{*}(\myState, a)
\end{equation}
If we can identify $Q^{*}$, we have an optimal policy to control the actions of the picker agent in the store environment in which he was trained. This policy can be determined using the referred Q-Learning approach, using the Bellman equation to approximate $Q^{*}$. The Bellman equation based on a Q-function can be expressed recursively by 
\def \LearningRate      { \alpha }
\def \myDiscountRate    { \gamma }
\begin{equation}
    Q(s, a) = R + \myDiscountRate \max\limits_{a'} Q(s', a')
\end{equation}
The Q-learning equation to iteratively approximate $Q^{*}$ is given by
\begin{equation}
    Q^{new}(s, a) = Q(s, a) + \LearningRate\Big(R + \myDiscountRate \max\limits_{a} Q(s', a) - Q(s,a) \Big)
\end{equation}

Where the learning rate $\LearningRate$ adjusts how different previous and new Q values are from each other. As the agent explores more and more the environment, the approximated $Q^{new}$ values will converge to $Q^{*}$. To extract an optimal picking policy for a store environment, we propose \myalgref{algo:q_learning}.

\begin{algorithm}[hbt]
\caption{Q-Learning procedure}\label{algo:q_learning}
\begin{algorithmic}[1]
\footnotesize{
\Procedure{QLearning($MDP$, $\LearningRate$, $\myDiscountRate$, $\epsilon$)}{}
\State Initialize $Q(s, a)$ using random optimistic values; \label{algo:ql-initialize_q}
\For{each episode} \label{algo:ql-foreach_episode}
    \State Initialize the environment with the initial state $s_0$ (picker at the depot); \label{algo:ql-initialize_env}
    \While{$\myState$ is not the final state (the store is not closed)} \label{algo:ql-not_final_state}
        \State Choose an action $\myAction$ available from the current state $\myState$ following an $\epsilon\text{-greedy}$ strategy;  \label{algo:ql-choose_action}
        \State Execute action $\myAction$, receive a reward $\myReward$, and observe a new state $s_{\myEpoch + 1}$;
        \State Update Q-function by doing $Q(\myState, \myAction) \gets Q(\myState, \myAction) + \LearningRate\Big(\myReward + \myDiscountRate \argmax\limits_{a} Q(s_{\myEpoch+1}, a) -  Q(\myState, \myAction) \Big) $ \label{algo:ql-recursive_bellman_update}
    \EndWhile
	\If {$Q$ converges}
    	\State Break;
    \EndIf
\EndFor
\State \Return Q
\EndProcedure
}
\end{algorithmic}
\end{algorithm}

The algorithm starts by initializing the Q-function with random optimistic values (\myalgref{algo:ql-initialize_q}, line \ref{algo:ql-initialize_q}). These optimistic values promote the exploration of new states. Then, a cyclic procedure is repeated for a certain number of episodes (\myalgref{algo:ql-initialize_q}, line \ref{algo:ql-foreach_episode}). First, the store environment initializes at the initial state $s_{0}$ (\myalgref{algo:ql-initialize_q}, line \ref{algo:ql-initialize_env}) and, while it does not reach the final state $s_{\myTimeHorizon}$ (\myalgref{algo:ql-initialize_q}, line \ref{algo:ql-not_final_state}), it chooses an action, executes it, and updates the Q-function (\myalgref{algo:ql-initialize_q}, lines \ref{algo:ql-choose_action} to \ref{algo:ql-recursive_bellman_update}). The choice of the action is performed using an $\epsilon$-greedy, meaning that a random action is chosen with probability $\epsilon$, and the action with the best Q value is chosen with probability $1-\epsilon$. After finishing an episode, the algorithm verifies if the Q-function converged (i.e., check if the maximum difference between old and new q-values is lower than a convergence threshold). If it did converge, the algorithm returns the Q-function, otherwise, it continues to the next episode.

\subsection{Illustrative example}
\label{subsection:illustrative_example}

\begin{figure}[t]
  \centering
  \includegraphics[width=0.65\textwidth]{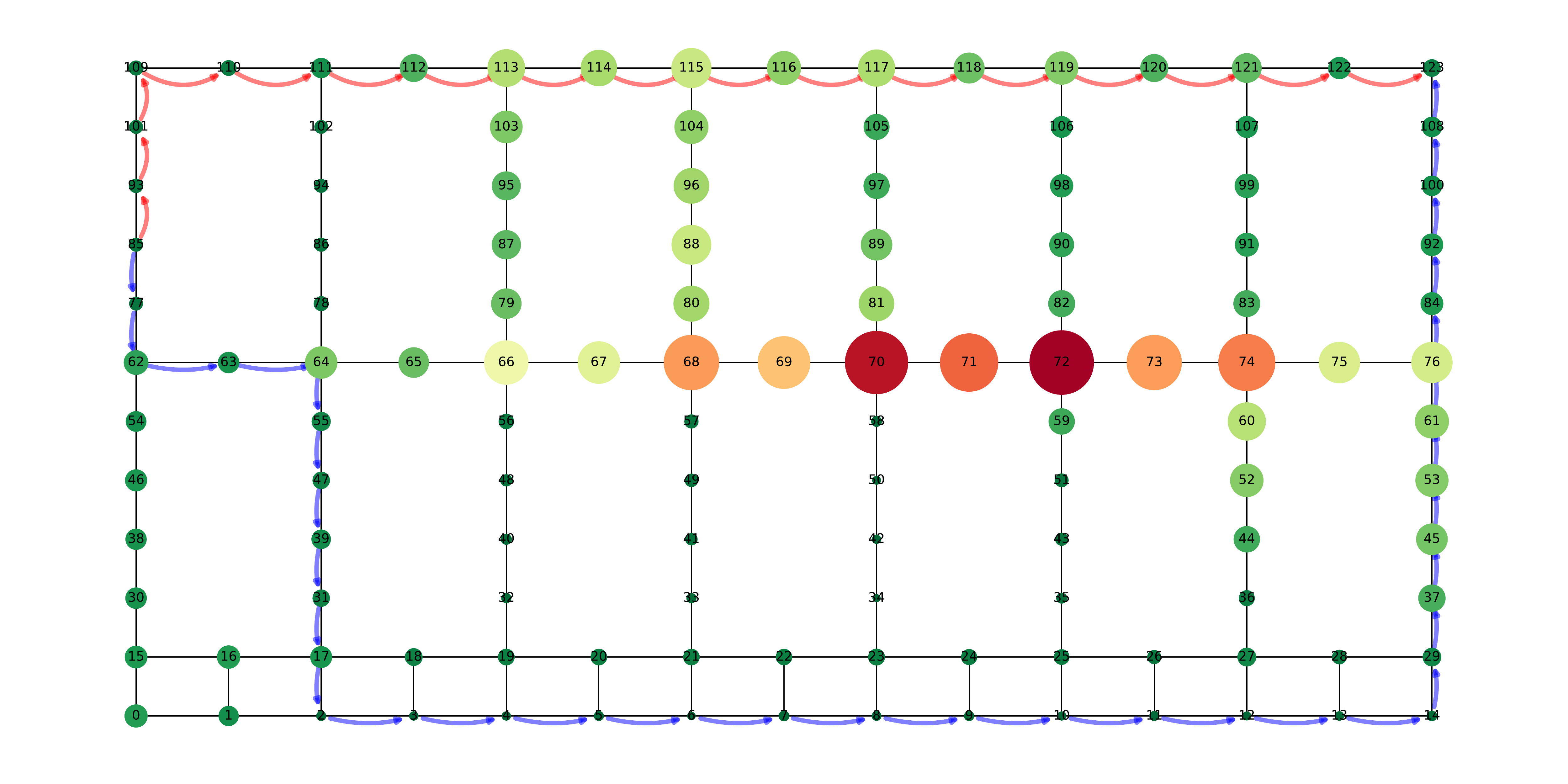}
  \caption{Illustrative example of a \textbf{SP} (red) and a \textbf{QL} path (blue).}
\label{fig:example_path}
\end{figure}

The solution approach developed in this work converges to optimal policies that are substantially different from a shortest path-based policy, depending on the reward shape assumed. \myfigref{fig:example_path} illustrates a comparison between two paths with similar origin and destination locations. In this figure, the size of the nodes measures the average number of customers that have been present at each location. We observe that the central aisle and the product positions in the top middle aisles were more crowded during the simulated episode. The \textbf{SP} policy (shortest path-based), in red, results in a shorter distance but it encounters more customers. On the other hand, the \textbf{QL} policy (Q-Learning-based), in blue, results in a large detour that, on average, encounters fewer customers.

\begin{figure}[hbt]
  \centering
\begin{adjustbox}{width=0.65\textwidth, trim=0cm 3cm 3.2cm 0cm, clip}{
  \includegraphics[width=1.0\textwidth]{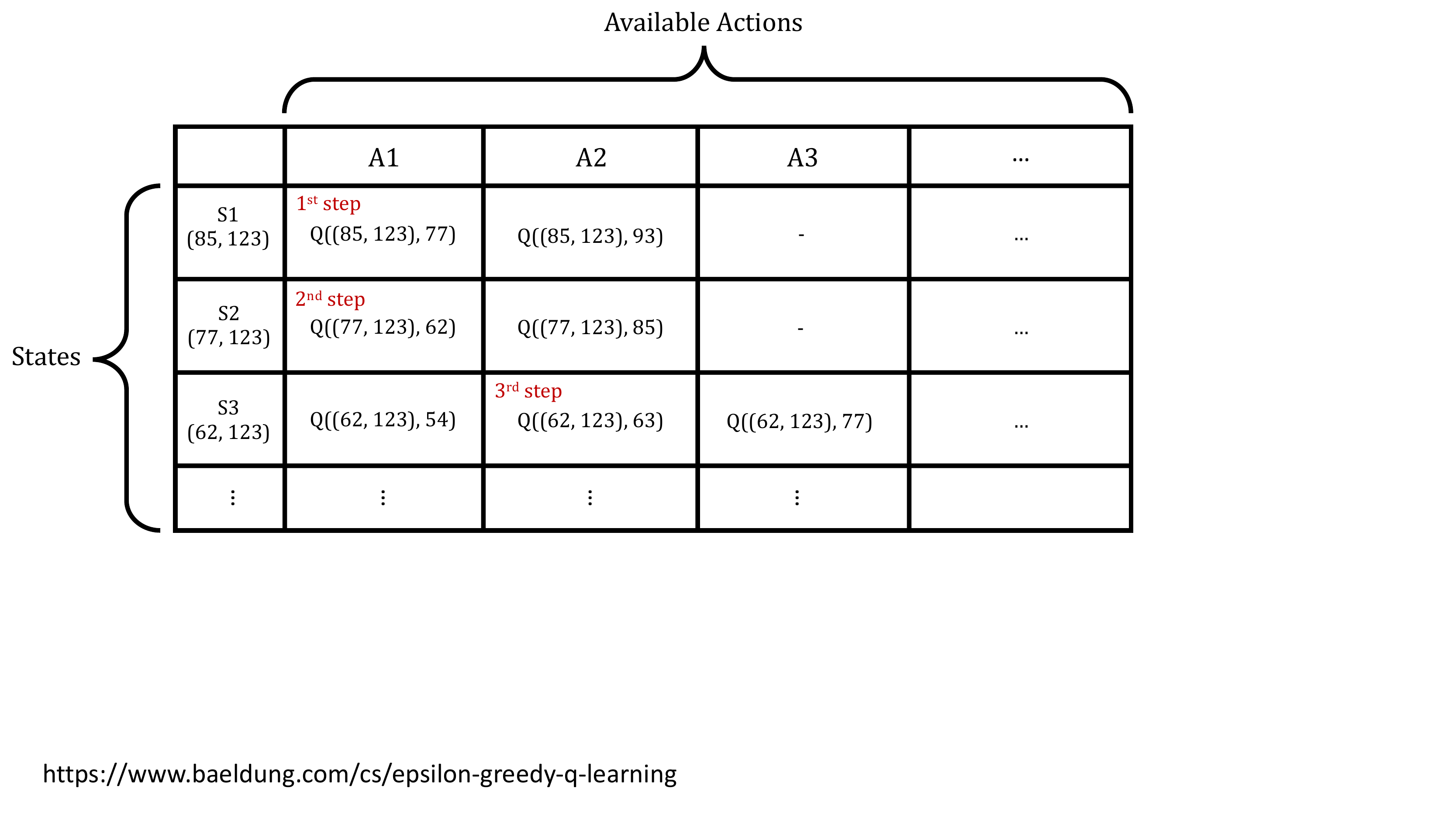}
}\end{adjustbox}
\caption{An example with three steps obtained from the Q-table that was used to obtain the Q-learning path in the illustrative example.}
\label{fig:qtable}%
\end{figure}

The blue path is obtained by simply following a policy extracted from the Q values. \myfigref{fig:qtable} presents an example of three steps performed using a Q-table (lookup table).

Each row represents a state that is a tuple containing the current position (node 85) and the target position (node 123). Each column corresponds to a possible action considering the current positions of the picker. For instance, in the first step, the picker could walk to node 77 or node 93. Since the Q value of node 77 is larger, the picker moves to position 77.

\section{Algorithm convergence}
\label{sec:computational_experiments}

In this section, we present the computational experiments to explore the efficacy and efficiency of the proposed approach on synthetic instances. Firstly, we describe the process that was used to generate the synthetic instance set. Secondly, we perform a search on the training parameters to find good combinations of learning rate, discount rate, and $\epsilon$-greediness and validate the convergence of the proposed approach on different problem sizes.

\subsection{Synthetic instances}
\label{subsec:synthetic_instances}
To validate the proposed approach, synthetic instances were generated by considering different simulation parameter combinations. We tested combinations of four store layouts with different sizes and with three different customer concentrations inside the store. As such, store environments can be tiny, small, medium, and large, and the products of customers' shopping lists can be more concentrated near the entrance, the middle, or the back of the store. The number of products in customers' shopping lists follows a uniform distribution with a minimum of $1$ and maximum of $10$ products. These products and their picking sequence are randomly chosen and customers follow the shortest path when moving from one product position to another product position. The arrival rate of physical customers is set to $2$ and the arrival rate of online orders is set to $0.2$. In each episode, the store is open for $8$ hours, and the maximum number of customers inside the store $\myStoreMaxPhysicalCustomers$ is set to $50$. The maximum number of customers per node $\myNodeMaxPhysicalCustomers$ is $5$ and each customer takes $30$ seconds to pick a product. The walking speed is set to $1$ m/s. \myfigref{tab:simulation_parameters} summarizes the sets of parameters that were used to build $12$ synthetic instances. Regarding rewards and penalties in the synthetic instances, when a picker is in the same node where a customer is, it is penalized by $3$ points. If a picker is seen by a customer (i.e., the customer is in the nodes accessible by the picker), the picker is penalized by $1$ point. Picker steps are penalized by $1$ point and products picked are rewarded with an amount of $100$ points.

\begin{table}[hbt]
    \centering
    \caption{Sets of parameters used to build the generated instances.}
\begin{adjustbox}{width=0.65\textwidth}
    \begin{tabular}{rl}
    \toprule
         Store layout                                & $\{\text{Tiny}, \text{Small}, \text{Medium}, \text{Large}\}$ \\
         Customer concentration                      & $\{\text{Entrance}, \text{Middle}, \text{Back} \}$  \\
         Physical customer arrival rate              & $\{2 \}$ \\
         Online order arrival rate                   & $\{0.2 \}$ \\
         Open time (h)                               & $\{8 \}$ \\
         Maximum number of customers inside          & $\{ 50 \}$ \\
         Maximum number of customers in a location   & $\{ 5 \}$ \\
         Product picking time (s)                    & $\{ 30 \}$ \\
         Walking speed (m/s)                         & $\{ 1 \}$ \\
         Customer encounter penalty (p1)      & $\{ 3   \}$ \\
         Customer visible penalty (p2)        & $\{ 1 \}$ \\
         Step penalty (p3)                    & $\{ 1 \}$ \\
         Picking reward (p4)                  & $\{ 100 \}$ \\
    \bottomrule
    \end{tabular}
    \label{tab:simulation_parameters}
\end{adjustbox}	
\end{table}

\subsection{Q-learning training and results}
\label{subsec:synthethic_instances_training_testing}
Despite the potential of reinforcement learning approaches, tuning them is often a non-trivial and time-consuming task. This is essentially because testing parameter combinations can be an expensive and noisy process. To overcome these difficulties, we tested several parameter configurations to determine policies for each synthetic instance. Each configuration was run for 5000 episodes using a single thread. All computations were run on Intel Xeon @ 2.5 GHz processing units and the mathematical solver adopted to compute picking sequences of the pickers was Gurobi 9.5. \mytabref{tab:training_parameters} presents the parameter sets that were used to build each training configuration.


\begin{table}[hbt]
    \centering
    \caption{Sets of reinforcement learning parameters used to build training configurations. }
\begin{adjustbox}{width=0.3\textwidth}
    \begin{tabular}{rl}
    \toprule
         Learning rate ($\LearningRate$)      & $\{ 0.95, 0.97, 0.99 \}$ \\
         Discount rate ($\myDiscountRate$)    & $\{ 0.50, 0.70, 0.90 \}$ \\
         $\epsilon$-greediness ($\epsilon$)   & $\{ 0.01 \}$ \\
    \bottomrule
    \end{tabular}
    \label{tab:training_parameters}
    \end{adjustbox}	
\end{table}
The cumulative rewards obtained for the training episodes are presented in \mytabref{tab:synthetic_training_and_testing_results}. 
The first two columns, ``Layout" and ``Paths", indicate the layout of the store and the type of shopping paths performed by the customers. The third column, ``Best Parameters", indicates the combination of parameters ($\LearningRate | \myDiscountRate | \epsilon$) that obtained the values presented in the respective row. Columns four to six present the minimum, average, and maximum cumulative rewards obtained across all the tested parameter configurations. The last column presents the coefficient of variation over the last $50$ observations.

\begin{table}[htb]										
\centering												
\caption{Cumulative rewards obtained in the training procedure (5000 episodes).}	
\begin{adjustbox}{width=0.6\textwidth}
\begin{tabular}{llcrrrc}				
\toprule
& & & \multicolumn{4}{c}{\textbf{Cumulative rewards}} \\ \cline{4-7} 
Layout & Paths & \makecell{Best Parameters \\ ($\LearningRate | \myDiscountRate | \epsilon$) } & Min. & Avg. & Max. & \makecell{CV \\ (last $50$ obs.)} \\ \midrule
tiny & 	far & 	$ 0.97\, | \,0.9\, | \,0.01 $ & 	28989.54 & 	29262.92 & 	29440.60 & 	0.11 \\ 
tiny & 	middle & 	$ 0.99\, | \,0.7\, | \,0.01 $ & 	27240.37 & 	28563.90 & 	29262.10 & 	0.11 \\ 
tiny & 	near & 	$ 0.95\, | \,0.9\, | \,0.01 $ & 	28325.40 & 	28810.05 & 	29073.35 & 	0.10 \\ 
small & 	far & 	$ 0.97\, | \,0.9\, | \,0.01 $ & 	23471.06 & 	26828.68 & 	28569.88 & 	0.11 \\ 
small & 	middle & 	$ 0.97\, | \,0.9\, | \,0.01 $ & 	8571.05 & 	21363.16 & 	28362.71 & 	0.10 \\ 
small & 	near & 	$ 0.97\, | \,0.9\, | \,0.01 $ & 	11729.02 & 	22830.18 & 	28236.43 & 	0.10 \\ 
medium & 	far & 	$ 0.99\, | \,0.9\, | \,0.01 $ & 	-249624.48 & 	-74201.81 & 	26741.11 & 	0.11 \\ 
medium & 	middle & 	$ 0.99\, | \,0.9\, | \,0.01 $ & 	-280708.50 & 	-136167.61 & 	26352.44 & 	0.12 \\ 
medium & 	near & 	$ 0.97\, | \,0.9\, | \,0.01 $ & 	-275092.86 & 	-144711.93 & 	25733.31 & 	0.08 \\ 
large & 	far & 	$ 0.95\, | \,0.9\, | \,0.01 $ & 	-282733.08 & 	-172079.58 & 	22584.38 & 	0.10 \\ 
large & 	middle & 	$ 0.95\, | \,0.9\, | \,0.01 $ & 	-286259.32 & 	-152702.76 & 	17371.42 & 	0.18 \\ 
large & 	near & 	$ 0.97\, | \,0.9\, | \,0.01 $ & 	-285686.24 & 	-180264.63 & 	21524.09 & 	0.13 \\ 
\bottomrule	
\end{tabular}%
\label{tab:synthetic_training_and_testing_results}%
\end{adjustbox}						
\end{table}%

\begin{figure}[H]
  \centering
\begin{adjustbox}{width=0.8\textwidth}{
  \includegraphics[width=1.0\textwidth]{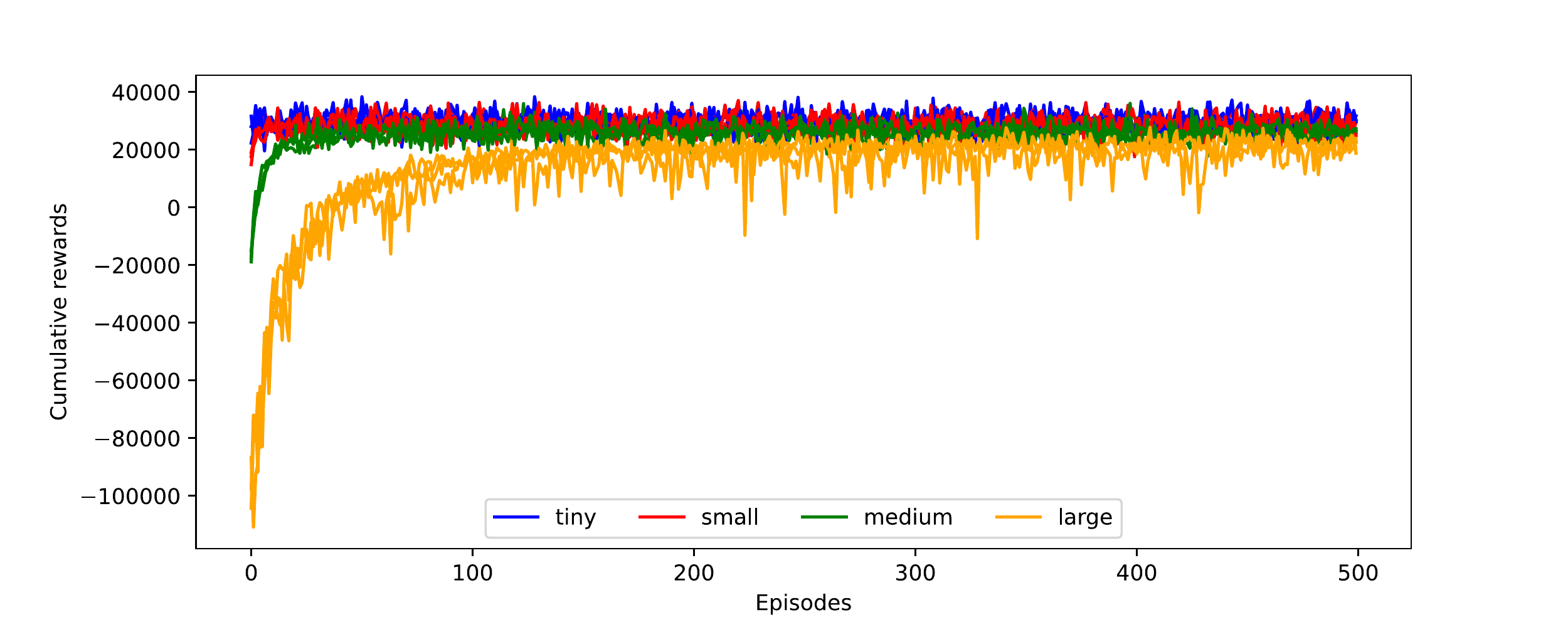}
}\end{adjustbox}
\caption{Cumulative rewards obtained by the best parameter configurations used in synthetic instances (first 500 episodes of the training session).}
\label{fig:synthetic_instances_convergence}%
\end{figure}

We observe that for the larger store layouts the difference between the minimum and maximum cumulative rewards obtained increases. As the problem size increases, the importance of choosing a good combination of parameters also increases. As expected, the cumulative reward obtained in larger stores is smaller, as the picker spends more time traveling through a larger number of nodes particularly when he/she needs to return to the preparation zone. Moreover, we observe that for the last $50$ observations the coefficient of variation of the rewards is low for all instances sizes, enforcing the idea that the algorithm converged in all of them. At this point, to better understand the algorithm convergence, we present \myfigref{fig:synthetic_instances_convergence}.

The cumulative rewards suggest that the algorithms converge in a few episodes. As expected, the size of the store considered in each instance is connected to the time it takes for the cumulative rewards to converge. While in instances with a medium layout we observe the algorithm converging after a few episodes, in instances with a large layout the cumulative rewards stabilize after more than $200$ episodes. Nonetheless, the proposed approach is successful in achieving picking policies that do not demand substantial computational power.


\section{Real-world application}
\label{sec:real_world_application}
This section details a real application in the context of a large European retailer. The devised approach was applied using the layout of a real-world store considering real shopping paths collected from customers using a mobile app that allows them to keep track of products they pick in-store as they are shopping. 

A very important aspect of the devised approach is that it does not require substantial computational power. The output of the Q-learning approach is a policy in the form of a Q-table (lookup table) that can be used in simple mobile devices (i.e., phones and PDAs). Moreover, this can be improved over time in an offline or in an online fashion, using new data collected from customer and picker shopping paths to update Q-tables. It is even possible to compute Q-tables tailored for different parts of the day or of the week, suited to the customer flows that are more likely to occur during these periods.

\subsection{Real-world instance}
\label{subsec:real_world_instance}

\begin{figure}[H]
  \centering
\begin{adjustbox}{width=0.8\textwidth, trim=1.1cm 4cm 0.8cm 3.7cm, clip}{
  \includegraphics[width=1.0\textwidth]{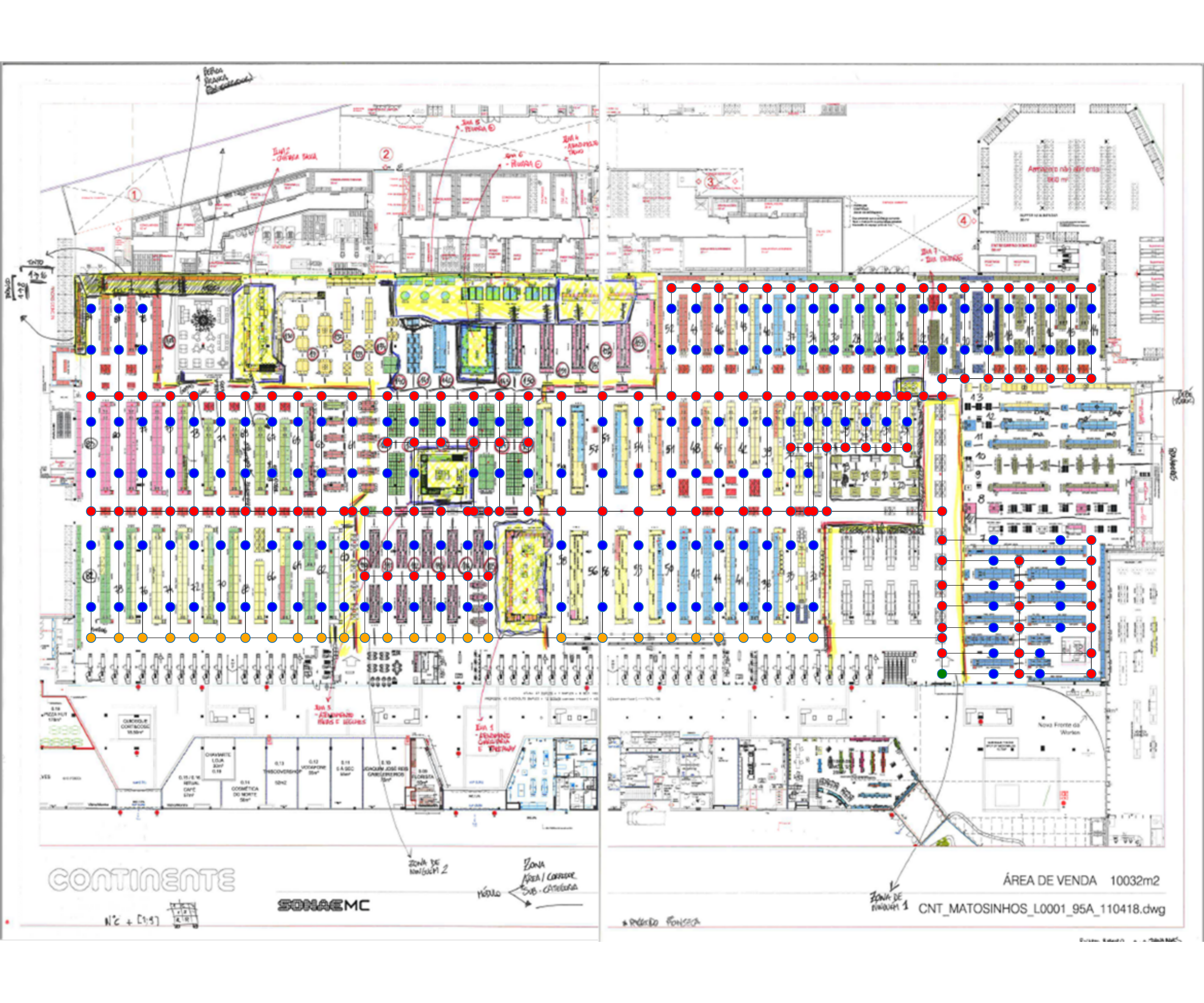}
}\end{adjustbox}
\caption{Representation of the store layout and graph considered in the real-world application.}
\label{fig:real_store_layout}%
\end{figure}

The first step to applying the devised approach to the considered real context was to represent the retail store with substantial detail. This is a particularly challenging process due to the lack of data regarding product positions, which occurs in practice. Large retailers are now digitizing physical stores as the opportunities provided by the new data collected are remarkable. However, in the retailer considered in this study, this process is still in its infancy. In the retail business, several products are likely to be relocated daily for a multitude of reasons (i.e., promotional activities, new products introduced in the assortment), which increases the difficulties of maintaining reliable information on product positions. Nonetheless, we mapped products to aisles, achieving a realistic representation of the real store. \myfigref{fig:real_store_layout} presents the store layout and graph considered in the real-world application. 

\def \realcustomerarrivalrate { 10 } 
\def \realorderarrivalrate { 0.3 } 
\def \realopentime { 8 } 
\def \realstoremaxcustomers { 300 }
\def \realnodemaxcustomers { 25 }
\def \realpickingtime { 30 } 
\def \realwalkingspeed { 1 } 

In this figure, the nodes represented in red are aisle intersections and the nodes represented in blue are product positions. We opted for considering at most two product nodes inside each aisle as it already provides a good indication of the position of each product. In the considered graph, customers are allowed to enter the store by the node represented in green and, after traversing their shopping path, they randomly choose one of the exit nodes represented in orange. The sequence of products picked by the physical customers is based on real-world data collected using the mobile app provided by the retailer. To compute the shortest path between each pair of products in the customer shopping path we resorted to an implementation of a Dijkstra algorithm. The remaining parameters were set according to discussions with the retail store managers. The arrival rate of physical customers $\myStoreCustomerRate$ is set to $\realcustomerarrivalrate$ and the arrival rate of online orders $\myOnlineOrderRate$ is set to $\realorderarrivalrate$.  The maximum number of customers allowed in the store is set to $\realstoremaxcustomers$ and we did not set a maximum number of customers per node in the real application. Each customer takes $\realpickingtime$ seconds to pick a product and walks at a speed of $\realwalkingspeed$ m/s. The store is open for $\realopentime$ hours. The picker agent is awarded with $50$ points for each product picked, it is penalized by $3$ points for each customer encountered in the same node, and it is penalized by $1$ for each customer seen in adjacent nodes. 

\begin{figure}
     \centering
     \begin{subfigure}[b]{0.49\textwidth}
         \centering
         \includegraphics[width=\textwidth]{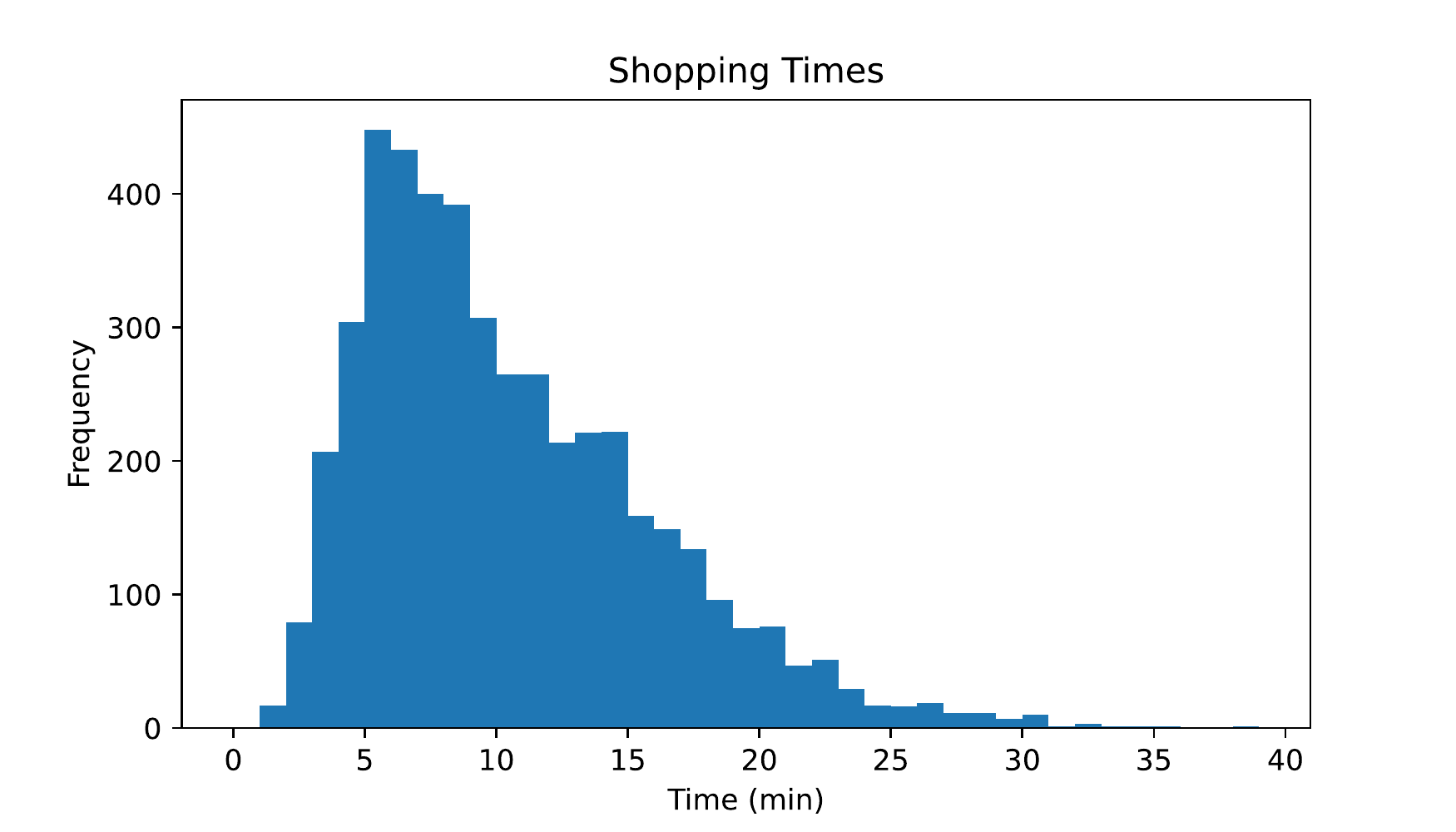}
         \subcaption{Shopping times histogram.}
         \label{fig:shopping_times}
     \end{subfigure}%
     \hfill
     \begin{subfigure}[b]{0.49\textwidth}
         \centering
         \begin{adjustbox}{width=1.0\textwidth, trim=0.5cm 1.9cm 0.5cm 1.5cm, clip}{
         \includegraphics[width=\textwidth]{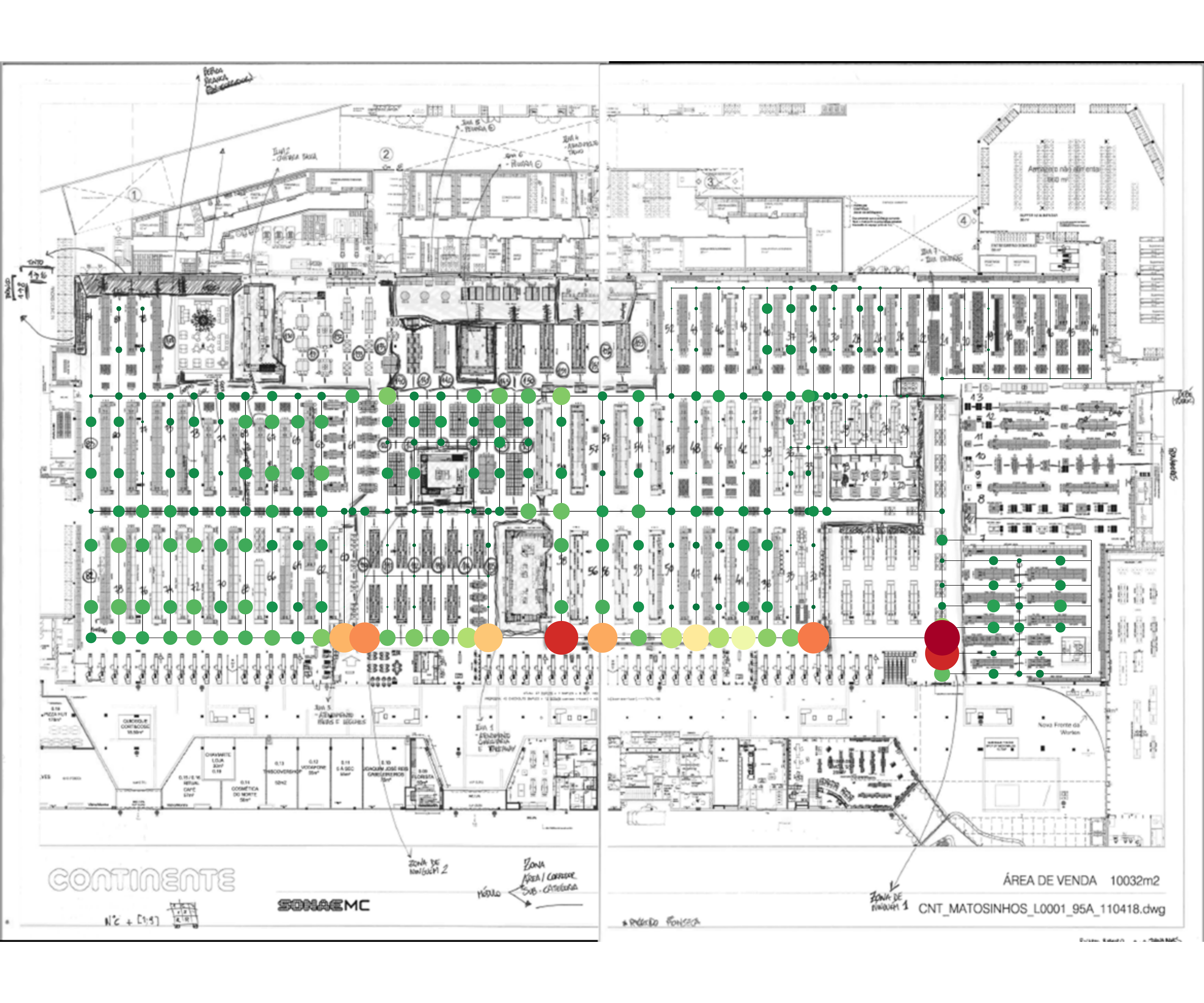}
         }\end{adjustbox}
         \subcaption{Most crowded nodes.}
         \label{fig:crowded_nodes}
     \end{subfigure}
     \hfill
     \begin{subfigure}[b]{1\textwidth}
         \centering
         \begin{adjustbox}{width=0.8\textwidth, trim=0cm 1.1cm 0cm 0.7cm, clip}{
         \includegraphics[width=\textwidth]{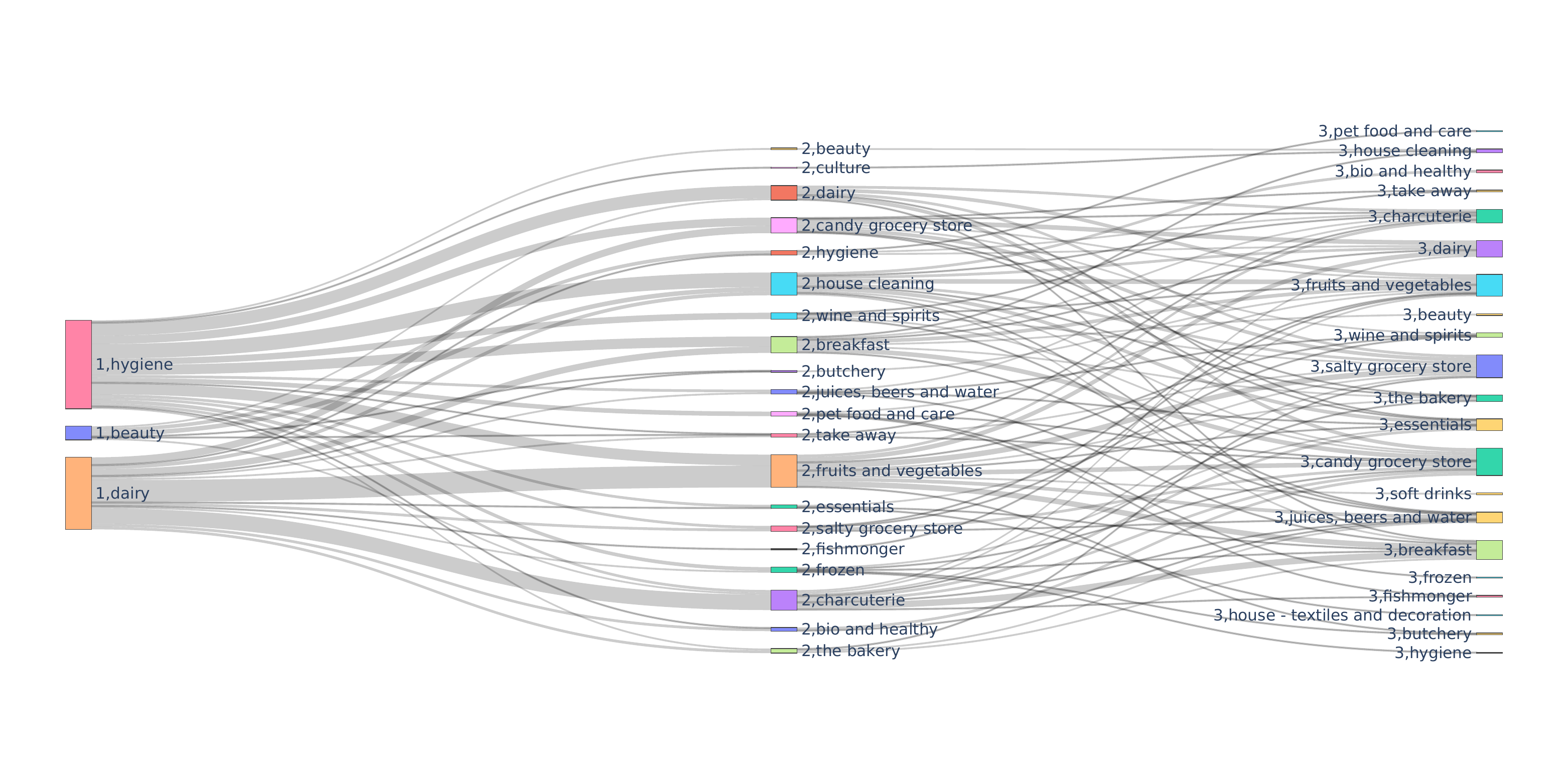}
         }\end{adjustbox}
         \subcaption{Categories visited by customers picking 3 products starting from hygiene, dairy, and beauty.}
         \label{fig:category_paths}
     \end{subfigure}
        \caption{Real-world store simulation environment metrics.}
        \label{fig:simulation_environment_metrics}
\end{figure}

To further describe the real-world store simulation environment that was used to train the picker agents, we present \myfigref{fig:simulation_environment_metrics} containing the distribution of the customer shopping times, a representation of the most crowded areas inside the store (average number of customers per decision epoch), and examples of category paths performed by customers picking only 3 products (for better visualization) and starting from the categories hygiene, beauty, and dairy.

We observe that most customers stay in-store for a period of $5$ to $15$ minutes. The most crowded nodes are the ones on the left side of the image, mostly corresponding to food products, and at the bottom of the image corresponding to the front aisle of the store near the cashiers. The order preparation zone from which the pickers depart is located at the top right of the picture. This area is usually less crowded as it includes work tools and various materials for DIY activities. We also observe that customers follow very distinct shopping paths even if they share a few nodes/products in their shopping lists.

\subsection{Training in a real-world context}
\label{fig:real_store_training_and_testing}%

The training process that was used in the real-world case was similar to the one presented in \mysecref{subsec:synthethic_instances_training_testing}. First, we set up several training sessions considering different combinations of reinforcement learning parameters, namely, learning rates $\LearningRate$, discount rates $\myDiscountRate$, and exploration probabilities $\epsilon$. The reinforcement learning parameter sets that were used are presented in  \myfigref{tab:real_training_parameters}.

\begin{table}[hbt]
    \centering
    \caption{Sets of reinforcement learning parameters to build training configurations for the real-world instances. }
\begin{adjustbox}{width=0.45\textwidth}
    \begin{tabular}{rl}
    \toprule
         Learning rate ($\LearningRate$)      & $\{0.1, 0.5, 0.9, 0.95, 0.97, 0.99 \}$ \\
         Discount rate ($\myDiscountRate$)    & $\{0.1, 0.5, 0.9, 0.95, 0.97, 0.99 \}$ \\
         $\epsilon$-greediness ($\epsilon$)   & $\{ 0.01, 0.03, 0.05, 0.07, 0.09 \}$ \\
    \bottomrule
    \end{tabular}
    \label{tab:real_training_parameters}
    \end{adjustbox}	
\end{table}

\begin{figure}[hbt]
  \centering
\begin{adjustbox}{width=0.8\textwidth}{
  \includegraphics[width=1.0\textwidth]{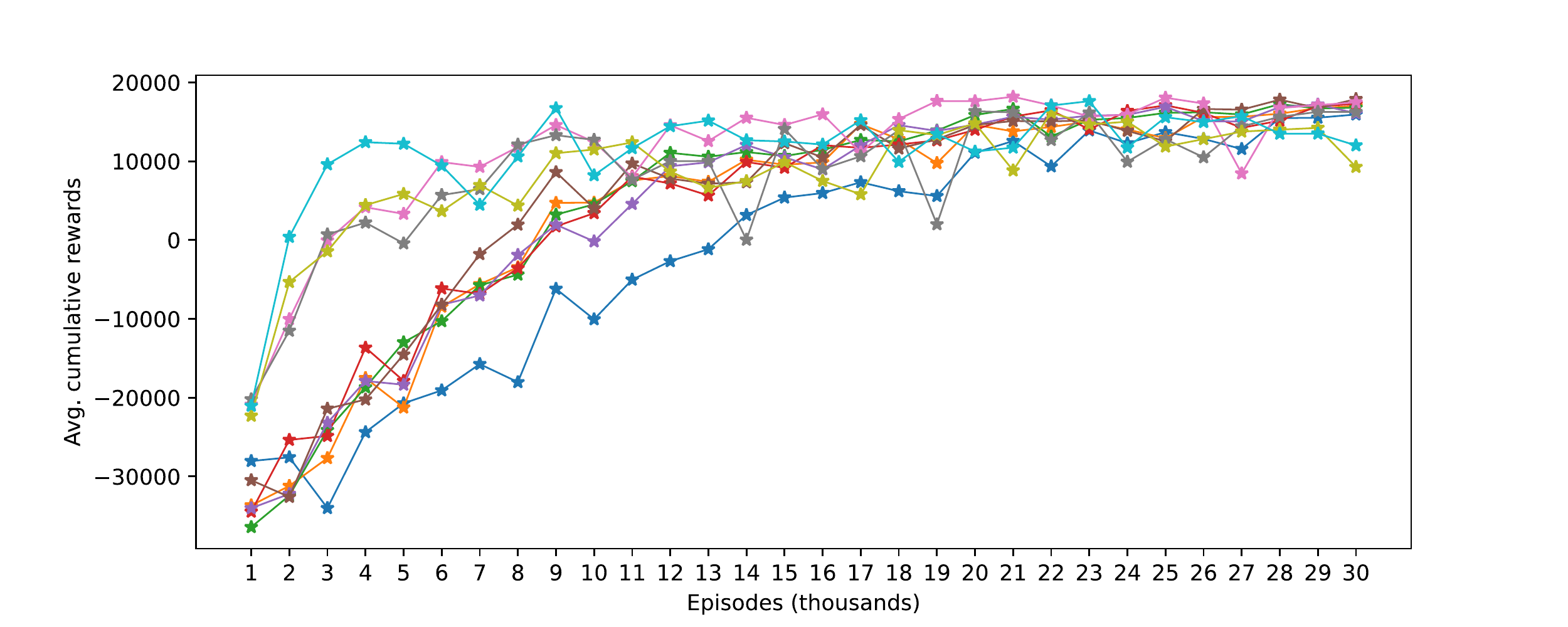}
}\end{adjustbox}
\caption{Average cumulative rewards for the top $10$ policies among all parameter configurations in the real-world store environment.}
\label{fig:real_instances_convergence}%
\end{figure}

A total of $30 000$ episodes were run for each configuration. Policies and the average cumulative rewards were saved every $1 000$ episodes. The $10$ best policies (in terms of average rewards) obtained in the training results related to the real retail store environment are presented in \myfigref{fig:real_instances_convergence}.
Every $1000$ episodes we restarted the computation of the average cumulative rewards. Note that after the first $1000$ episodes some policies were already performing better but had not converged. In general terms, the best policies converged after $20000$ episodes in the training procedure. A few policies could not match the performance of the best policies even after $30000$ episodes. 

In some reinforcement learning problems, average cumulative rewards obtained during the training phase (considering random actions) can be different from the ones obtained during the testing phase (constantly choosing the optimal action according to the Q-table). There is a large debate on how to evaluate these approaches and whether the $\epsilon$-greedy exploration should be kept during production or testing phase where no updates will be made to the picking policy. In some problems, it may make sense to keep this exploration feature as the agent may get stuck in the presence of unseen states. This is done successfully in \citet{mnih2015humanlevel} while playing Atari games. When solving path problems, non-optimal policies can be affected by cycles in case there is no randomness considered. For this reason, we consider that the $\epsilon$ parameter is to be maintained when the policy is used in practice.




\subsection{Managerial insights}
\label{subsec:managerial_insights}

The results shown in the previous sections suggest that the method can learn picking policies to be applied in real retail store environments. Indeed, we can observe the picker learning as he/she achieves higher and higher cumulative rewards. Nonetheless, one may ask whether we need all this work to pick a set of products or if it would be sufficient to just follow simpler policies as it is the case of a shortest path policy uniquely based on distance metrics.

The main questions that motivated us to solve this problem still remain. Can we improve customer experience? What is the impact on the number of picked orders? 

To answer these questions, we measure the performance of four different picking policies regarding business-related indicators such as the number of orders/products picked and customer encounters. 

\setlist[description]{font=\normalfont\space, leftmargin=0pt}
\begin{description}
       \item[\textbf{Q-learning (QL)}] Policy trained using the devised Q-learning that aims at finding a good compromise for picking orders while avoiding customer encounters.
    \item[\textbf{Shortest Path (SP)}] Policy based on a shortest path policy (employing a Dijkstra algorithm) aiming at picking the most orders and ignoring customer encounters. Each step is obtained by solving a shortest path problem considering arcs weighted by their length. 
    \item[\textbf{Myopic Policy (MP)}] Policy based on a myopic behavior, which aims at following the nodes with least customers, without getting farther from the target picking position.
    \item[\textbf{Crowded Nodes (CN)}] Policy that minimizes the number of customer encounters based on a static measure of the average of customers in each node. Each step is obtained by solving a shortest path problem considering arcs weighted by the average number of customers traversing them.
\end{description}

Additionally, for each policy, we considered two types of routing costs to sequence the picking order of the products of an order. Therefore, whenever an online order arrives, the \gls{srp} can be solved using a cost matrix considering distances or a crowdedness measure (average number of customers between the origin and destination nodes). These two routing basis are called \textbf{Arc Distance} and \textbf{Arc Crowdedness}, respectively. This will affect the sequence in which the products of an order are picked.


\mytabref{tab:real_business_indicator_results} presents the indicators obtained for the four policies using each of the routing basis considered.

\begin{table}[H]							
\centering	
\caption{Average indicators obtained for 500 episodes of the \\ Shortest Path (SP), Crowded Node (CN), Myopic Policy (MP), and Q-learning (QL) policies.}
\begin{adjustbox}{width=0.7\textwidth}
\begin{tabular}{rrrrrr}
\toprule 
Routing Basis & Policy & \makecell{Avg. \\ Rewards} & \makecell{Avg. \\ \#Orders} & \makecell{Avg. \\ \#Products} & \makecell{Avg. \\ \#Encounters} \\ \midrule
 Arc Distance &  QL & 16582.12 & 41.03 & 487.01 & 1025.39 \\ 
 Arc Distance & SP & 15388.14 & 46.56 & 553.79 & 2072.22 \\ 
 Arc Distance &  CN & 16085.83 & 38.51 & 456.64 & 745.71 \\ 
 Arc Distance &  MP & 15294.34 & 47.65 & 564.76 & 2134.58 \\ \cline{2-6}
               & Avg. & 15838.69 & 43.45 & 515.55 & 1494.40 \\
\midrule
 Arc Crowdedness &  QL & 16997.61 & 41.49 & 491.58 & 1018.92 \\ 
 Arc Crowdedness & SP & 14991.66 & 44.45 & 527.50 & 1868.61 \\ 
 Arc Crowdedness &  CN & 16845.25 & 39.84 & 470.69 & 768.99 \\ 
 Arc Crowdedness &  MP & 15244.25 & 45.66 & 542.28 & 1891.15 \\ \cline{2-6}
               & Avg. & 16019.69 & 42.86 & 508.01 & 1386.92 \\
 \bottomrule 
\end{tabular}%
\label{tab:real_business_indicator_results}%
\end{adjustbox}						
\end{table}%

Analyzing the case where route sequences are obtained using arc distances, we observe that \textbf{SP} and \textbf{MP} are the most operationally efficient policies, picking an average of $46.56$ and $47.65$ orders, respectively. Policies \textbf{QL} and \textbf{CN} are not so efficient in terms of number of orders picked, picking $41.03$ and $38.51$ orders, respectively. These policies are clearly focused on improving customer experience, lowering the average number of customer encounters by more than $50\%$ compared to policies \textbf{SP} and \textbf{MP}. Policy \textbf{CN} achieves the lowest number of customer encounters, yet it could pick the lowest number of orders among all policies. Policy \textbf{QL} drastically reduced the number of customer encounters without heavily compromising the number of orders picked.

Observing the results obtained when route sequences are obtained using the arc crowdedness measure, we observe a similar behavior of the policies, in relative terms, regarding operational efficiency and customer experience. However, the route sequences obtained suffer from a reduction in the average number of orders picked, reducing by $1.35\%$, from an average $43.45$ orders to $42.86$. The number of customer encounters slightly improved, reducing by $7.19\%$, from an average of $1494.40$ encounters to $1386.92$. Therefore, using arc crowdedness as a routing basis helped to find small improvements in customer experience for all policies at a cost of a small reduction in the number of orders picked.

We observe the most operationally efficient policies are the policies based on following shortest distance paths (\textbf{SP} and \textbf{MP}). In terms of providing the best customer experience, the \textbf{CN} policy achieves the lowest average number of customer encounters. However, these policies are totally focused on one objective only. While the \textbf{SP} and \textbf{MP} policies cannot avoid customers, the \textbf{CN} policy is not aware that the ultimate objective is to pick orders. The \textbf{QL} policy achieves a good compromise in picking a large average number of orders of and encountering a low average number of customers.

An interesting detail on these results is that the \textbf{QL} had to be re-trained when the routing basis was changed from distance to crowdedness. Note that we are dealing with a problem that suffers from the ``curse of dimensionality'' in the number of states. Given that the route sequences were changed, different pairs of current position and target position (state) started to appear more frequently. Since the \textbf{QL} policy trained with arc distances had not visited these states sufficiently, the policy had to be trained in an environment where the route sequences considered the arc crowdedness.



\begin{figure}[hbt]
  \centering
    \begin{adjustbox}{width=0.8\textwidth, trim=2.8cm 4.1cm 2.7cm 3.7cm, clip}{
      \includegraphics[width=1.0\textwidth]{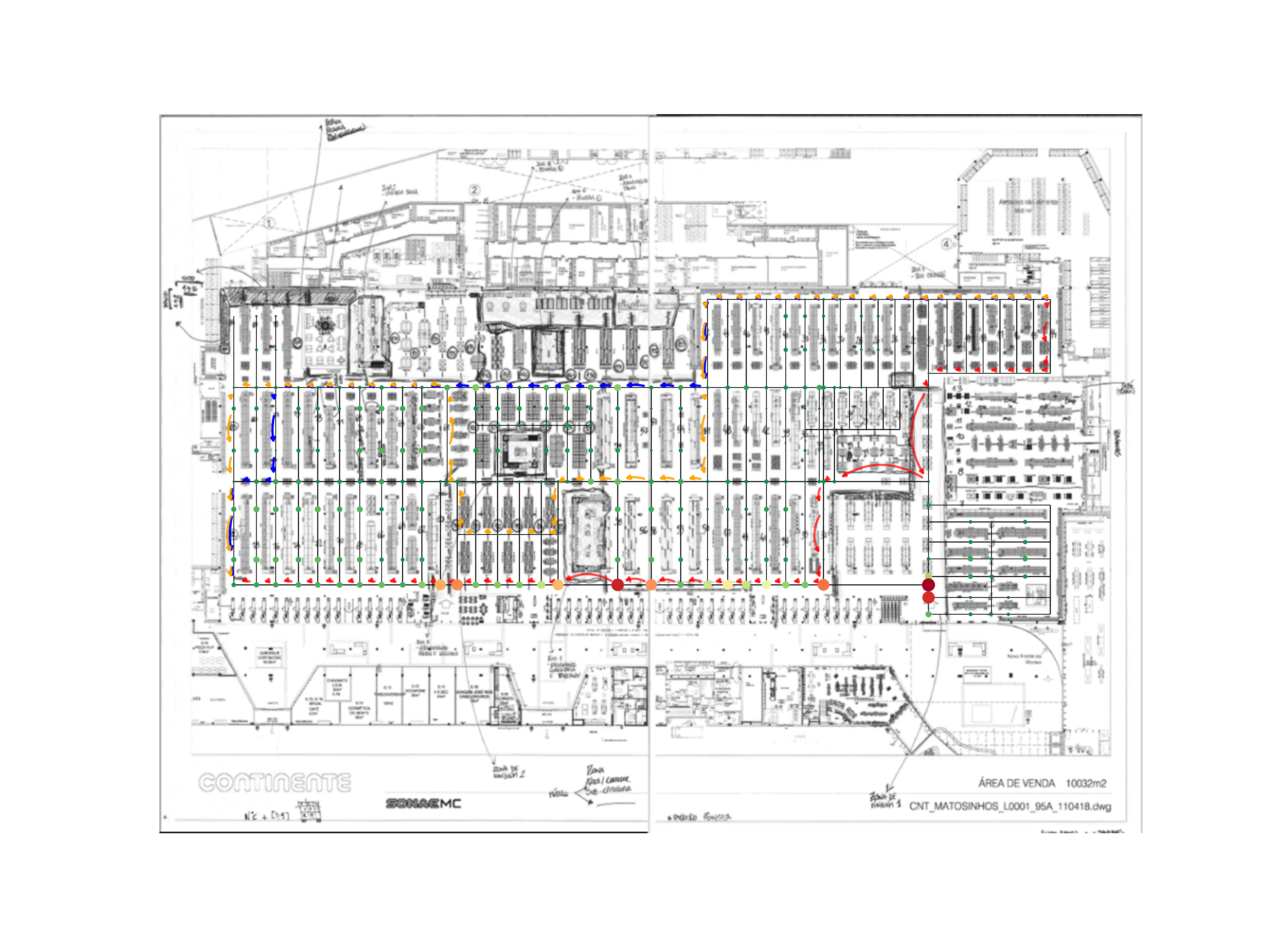}
    }\end{adjustbox}
  \caption{Illustrative example of a \textbf{SP} path (red), a \textbf{CN} path (orange), and a \textbf{QL} path (blue).}
  \label{fig:real_paths_ql_vs_spp}%
\end{figure}

The interesting behavior learned by the picker agent can be observed in several origin-destination pairs obtained in the policy. In \myfigref{fig:real_paths_ql_vs_spp} we present an illustrative example of the paths followed by the \textbf{SP}, the \textbf{CN} and the \textbf{QL} policies. 

Although the \textbf{SP} policy performs fewer steps (traveling a shorter distance), it ends up visiting the most crowded nodes. On the other hand, the \textbf{QL} policy follows a longer path but it tries not to visit crowded nodes, resulting in fewer customer encounters. The \textbf{CN} policy performs the longest path as it is trying to avoid customers at all cost.

\section{Conclusion}
\label{sec:conclusions}

This paper proposes a new problem in online retail, the \gls{diprp}, and solves it through a reinforcement learning approach. This problem has gained particular relevance during the pandemic context of 2020, in which most retailers offering online shopping services resorted to new picking operations in-store. These new picking operations will not be ephemeral, they will prevail as an important fulfillment method among omnichannel retail players.

Our solution approach encompasses several techniques, integrating the fields of mathematical programming, simulation, and machine learning, to obtain an in-store picking policy that is competitive in terms of the number of picked products while minimizing the number of customer encounters. Our study provides evidence that reinforcement learning can be an important method for effectively solving new management problems.

The devised solution approach trains picker agents to perform shopping paths that reduce the disturbance of in-store customers during their shopping experience. Computational experiments performed both on synthetic instances and a real-world instance from a large European retailer show that the picker agents can learn efficient picking policies in a reasonable number of episodes. An important characteristic inherent to our approach is the low computational power demanded and the very short computational time required to obtain a decision.

In addition, based on the real-world case study, we provide interesting numeric results and solution visualizations to support a few managerial insights. As expected, the shopping sequences followed by the customers often result in crowded store areas. Due to the dynamic and unpredictable movement of customers in-store, these crowded store areas are difficult to predict. However, we observe that the devised picking policies avoid these areas to reduce the number of customer encounters. This is not the behavior of the shortest path-based policies, which, logically, can pick a few more products, but nearly double the number of customers disturbed in our simulations. Our approach, based on reinforcement learning, allows better control over the trade-off between operational efficiency and customer experience. Such result may empower retailers to further pursue their omnichannel fulfillment strategy.

Many developments can be further introduced in this problem. For instance, developing collaborative picking policies to coordinate picking teams in this problem would be an interesting exercise, particularly if the customer perception regarding the number of picker encounters can be assessed. 
Additionally, improvements to the simulation environment can be introduced to consider, for example, shopping cart capacities and slowdowns experienced in crowded aisles. 
Furthermore, exploring richer state variables may improve the quality of the picking policies. For instance, the number of customers encountered in each aisle can further influence the decision of the picker in an online fashion (probably, the state space would suffer from the so-called ``curse of dimensionality'' more severely in this case). 
Lastly, we would like to stress that there are multiple applications for the approach proposed in this paper. The future reserves the automation of several activities, such as robots for cleaning public spaces, robots for picking products at warehouses and stores, and autonomous vehicles. These technologies will likely be placed in narrow spaces that are shared with humans. Since these automations are typically not pleasant to the eye, our approach can be an interesting alternative in the future.


\ACKNOWLEDGMENT{%
The research leading to these results has received funding from the European Union's Horizon 2020 - The EU Framework Programme for Research and Innovation 2014-2020, under grant agreement 952060.
}

%
%
%


\bibliographystyle{informs2014} 
\bibliography{biblia} 


\end{document}